\documentclass{article}

 \usepackage[preprint]{neurips_2026}

\usepackage[utf8]{inputenc} %
\usepackage[T1]{fontenc}    %
\usepackage{hyperref}       %
\usepackage{url}            %
\usepackage{booktabs}       %
\usepackage{amsfonts}       %
\usepackage{amsmath}        %
\usepackage{nicefrac}       %
\usepackage{microtype}      %
\usepackage{xcolor}         %
\usepackage{graphicx}       %
\usepackage{capt-of}        %
\usepackage{amssymb}        %
\usepackage{algorithm}      %
\usepackage{algpseudocode}  %

\title{Lite3R: A Model-Agnostic Framework for Efficient Feed-Forward 3D Reconstruction}

\author{
Haoyu Zhang$^{1*}$\quad
Zeyu Zhang$^{1*\dag}$\quad
Zedong Zhou$^{1}$\quad
Yang Zhao$^{2}$\quad
Hao Tang$^{1\ddag}$\\ 
[0.3em]
$^1$Peking University\quad
$^2$La Trobe University\\
[0.1em]
\footnotesize $^*$Equal contribution.
$^\dag$Project lead.
$^\ddag$Corresponding author: bjdxtanghao@gmail.com.
}

\begin{document}

\maketitle

\begin{center}
  \begin{minipage}{0.9\linewidth}
    \centering
    \includegraphics[width=\linewidth]{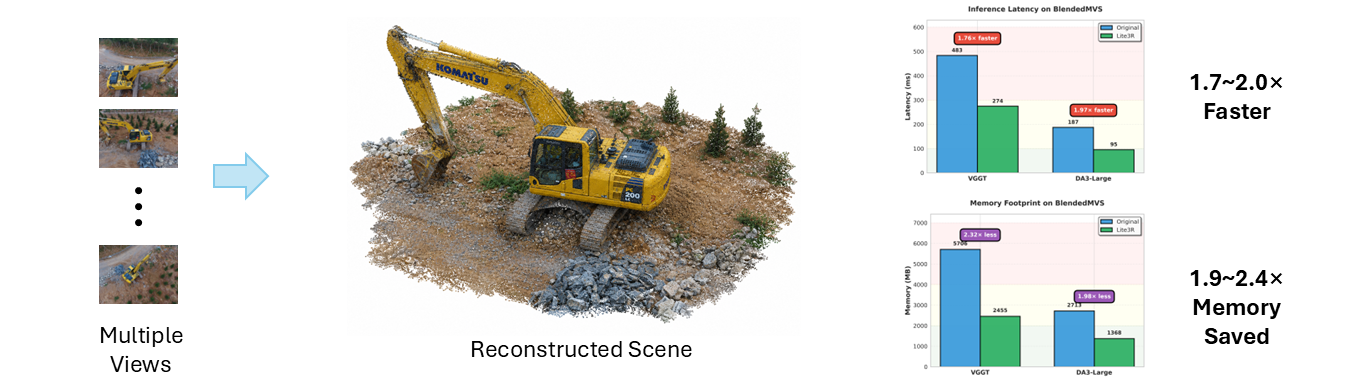}
    \captionof{figure}{Overview of Lite3R. A dense pretrained 3D reconstruction backbone is adapted into a lightweight student via Sparse Linear Attention, FP8-aware QAT, and partial attention distillation, improving deployment efficiency while preserving competitive geometry quality.}
    \label{fig:overview}
  \end{minipage}
\end{center}

\begin{abstract}
  Transformer-based 3D reconstruction has emerged as a powerful paradigm for recovering geometry and appearance from multi-view observations, offering strong performance across challenging visual conditions. As these models scale to larger backbones and higher-resolution inputs, improving their efficiency becomes increasingly important for practical deployment. However, modern 3D transformer pipelines face two coupled challenges: dense multi-view attention creates substantial token-mixing overhead, and low-precision execution can destabilize geometry-sensitive representations and degrade depth, pose, and 3D consistency. To address the first challenge, we propose \textbf{Lite3R}, a model-agnostic teacher--student framework that replaces dense attention with Sparse Linear Attention to preserve important geometric interactions while reducing attention cost. To address the second challenge, we introduce a parameter-efficient FP8-aware quantization-aware training (FP8-aware QAT) strategy with partial attention distillation, which freezes the vast majority of pretrained backbone parameters and trains only lightweight linear-branch projection layers, enabling stable low-precision deployment while retaining pretrained geometric priors. We further evaluate Lite3R on two representative backbones, VGGT and DA3-Large, over BlendedMVS and DTU64, showing that it substantially reduces latency (1.7--2.0$\times$) and memory usage (1.9--2.4$\times$) while preserving competitive reconstruction quality overall. These results demonstrate that Lite3R provides an effective algorithm--system co-design approach for practical transformer-based 3D reconstruction.
  Code:~\url{https://github.com/AIGeeksGroup/Lite3R}.
Website:~\url{https://aigeeksgroup.github.io/Lite3R}.
\end{abstract}
\section{Introduction}

Transformer-based 3D reconstruction has emerged as a powerful paradigm for recovering geometry and appearance from multi-view observations, offering strong performance across challenging visual conditions. Recent geometry-grounded pretrained models such as VGGSfM, DUSt3R, MASt3R, VGGT, and Depth Anything 3 have demonstrated notable gains in depth estimation, camera pose prediction, and holistic 3D consistency by leveraging dense multi-view attention and large-scale pretraining~\cite{wang2023vggsfm,wang2024dust3r,leroy2024mast3r,wang2025vggt,lin2025depthanything3}. As these models scale toward larger backbones and higher-resolution inputs, improving their efficiency becomes increasingly important for practical deployment. However, modern 3D transformer pipelines face two significant challenges: \textbf{(1)} dense multi-view attention creates substantial token-mixing overhead and memory pressure, making deployment costly~\cite{vaswani2017attention,wang2025vggt,lin2025depthanything3}; \textbf{(2)} low-precision execution can destabilize geometry-sensitive representations, as numerical perturbations propagate through multi-view matching and camera estimation, degrading depth, pose, and 3D consistency~\cite{micikevicius2022fp8,jacob2018quantization}.

To address these challenges, we identify two key motivations for designing an efficient 3D reconstruction system. First, to reduce the computational cost of dense attention without disproportionately degrading reconstruction quality, the lightweight model should retain important cross-view interactions through a structured sparsity mechanism rather than naive pruning or uniform compression~\cite{choromanski2021performer,wang2020linformer,zhang2025sla}. Second, to enable practical low-precision deployment, the system should incorporate quantization-aware training that accounts for the coupled effects of architectural modification and numerical perturbation under realistic hardware constraints~\cite{micikevicius2022fp8,jacob2018quantization}. 

Motivated by these observations, we propose \textbf{Lite3R}, a model-agnostic framework for efficient feed-forward 3D reconstruction. \textbf{(1)} Lite3R follows a teacher--student framework and replaces dense attention with \emph{Sparse Linear Attention} (SLA), which retains important cross-view interactions while substantially reducing attention cost and memory footprint. \textbf{(2)} We introduce a parameter-efficient FP8-aware quantization-aware training (FP8-aware QAT) strategy with partial attention distillation. Unlike conventional QAT that fine-tunes all parameters, our method freezes most pretrained backbone parameters and trains only lightweight linear-branch projection layers, thereby providing a lightweight adaptation path for low-precision deployment. To the best of our knowledge, this is among the first attempts to systematically bring FP8-aware QAT into transformer-based 3D reconstruction. \textbf{(3)} We conduct comprehensive experiments on two representative backbones, VGGT and DA3-Large, over BlendedMVS and DTU64 datasets, demonstrating that Lite3R substantially reduces latency (1.7--2.0$\times$) and memory footprint (1.9--2.4$\times$) while maintaining competitive depth, pose, and 3D reconstruction quality overall.

In summary, the contributions of our paper can be summarized in three folds:
\begin{itemize}
    \item We propose \textbf{Lite3R}, a model-agnostic teacher--student framework that replaces dense attention with Sparse Linear Attention to reduce computational cost while retaining useful cross-view interactions.
    \item We introduce a parameter-efficient FP8-aware QAT strategy with partial attention distillation, which freezes most pretrained parameters and trains only lightweight linear-branch projection layers, enabling low-precision deployment with a lightweight adaptation path.
    \item We conduct experiments on two representative backbones, VGGT and Depth Anything 3 Large (DA3-Large), over BlendedMVS and DTU64. The results show that Lite3R substantially reduces latency and memory footprint while maintaining a strong quality--efficiency tradeoff for practical deployment.
\end{itemize}

\section{Related Work}

\paragraph{Transformer-based 3D reconstruction.}
Recent 3D reconstruction systems increasingly rely on transformer backbones to aggregate information across multiple views and long token sequences. This improves global reasoning and cross-view correspondence, but also raises the cost of geometry inference relative to earlier local or convolution-dominated pipelines~\cite{schonberger2016colmap,pan2024glomap,yao2018mvsnet,chen2019pointmvsnet,vats2024gcmvsnet,zhang2023ramvsnet,liao2022wtmvsnet,yuan2024sdmvs,chen2024mvsplat}. Strong performance often depends on dense pretrained backbones such as DUSt3R, MASt3R, VGGSfM, VGGT, and Depth Anything 3, built on broader pretrained visual representations~\cite{oquab2023dinov2,dosovitskiy2020vit,ranftl2021dpt,yang2024depthanythingv2}, whose attention and linear layers dominate memory and latency~\cite{wang2024dust3r,leroy2024mast3r,wang2023vggsfm,wang2025vggt,lin2025depthanything3}. Recent efforts have also started to improve the efficiency of these geometry transformers more directly, for example through sparse/global attention redesigns for VGGT and feed-forward sparse 3D reconstruction variants~\cite{sparsevggt2026,fastvggt2025,flashvggt2025,speed3r2026}. Related systems such as MASt3R-SLAM, MASt3R-SfM, MV-DUSt3R+, Fast3R, Stream3R, TEST3R, and HAMSt3R push these backbones toward practical reconstruction, localization, and test-time adaptation~\cite{murai2025mast3rslam,duisterhof2025mast3rsfm,tang2024mv,yang2025fast3r,lan2026stream3r,test3r2025,rojas2025hamst3r}. Our work therefore focuses on \emph{adapting} strong geometry-grounded transformer backbones rather than replacing them.

\paragraph{Efficient attention for long-context geometry reasoning.}
A common route to improving transformer efficiency is to approximate dense attention with sparse, linear, or hybrid variants~\cite{katharopoulos2020transformers,choromanski2021performer,wang2020linformer,dao2022flashattention,shah2024flashattention3,zhang2025sla}. This design space is also beginning to appear in 3D geometry transformers, including block-sparse and descriptor-compressed variants tailored to VGGT-style architectures~\cite{sparsevggt2026,flashvggt2025}. For multi-view geometry, however, the challenge is not only to reduce complexity but also to retain the token interactions that carry cross-view correspondence cues. Purely linear approximations can therefore be brittle, while dense attention remains too expensive for deployment. Lite3R adopts a hybrid perspective: Sparse Linear Attention uses a sparse branch to retain high-value interactions and a lightweight linear branch to provide low-cost global context.

\paragraph{Low-precision adaptation of pretrained geometry models.}
Quantization is an appealing way to reduce the cost of large transformer models, yet geometry-sensitive models are vulnerable to numerical error because small perturbations can accumulate across long feature streams and degrade depth, pose, and 3D consistency~\cite{micikevicius2022fp8,jacob2018quantization}. More broadly, efficient vision-model deployment has explored data-efficient distillation, compact backbones such as DeiT, TinyViT, and MobileViT, and post-training quantization recipes such as SmoothQuant, AWQ, and GPTQ~\cite{touvron2021deit,wu2022tinyvit,mehta2022mobilevit,xiao2023smoothquant,lin2024awq,frantar2023gptq}. Directly converting a pretrained dense backbone to low precision is therefore often insufficient. We instead use a teacher--student framework in which structural lightweighting and low-precision robustness are learned jointly. Our FP8-aware QAT and partial attention distillation treat low precision as part of the adaptation process rather than a final conversion step~\cite{hinton2015distilling,zagoruyko2017at}.

\paragraph{System-oriented efficiency for end-to-end deployment.}
Recent work on efficient model serving has emphasized that kernel-level acceleration alone does not guarantee practical end-to-end gains; deployment also depends on memory traffic, activation storage, and execution scheduling~\cite{dao2022flashattention,shah2024flashattention3,xiao2023smoothquant,lin2024awq,frantar2023gptq}. This issue is especially pronounced in multi-view 3D reconstruction, where long sequences and large feature maps create heavy pressure on VRAM and bandwidth. It is also reflected in adjacent paradigms such as 3D Gaussian Splatting, DiViNeT, UniSDF, SERES, and recent bottleneck-aware 3DGS compression methods~\cite{kerbl20233dgaussians,vora2023divinet,wang2024unisdf,xu2025seres,zpressor2026}. Accordingly, we study latency, memory, and reconstruction quality together rather than isolated operator savings. This motivates Lite3R as an algorithm--system co-design approach in which attention replacement, FP8-aware QAT, and deployment efficiency work together.

\section{Method}

\subsection{Overview}

Lite3R follows a teacher--student framework for efficient geometry inference, with \emph{FP8-aware adaptation} as its main contribution. Starting from a dense pretrained geometry backbone, we build a lite student by replacing attention modules with Sparse Linear Attention (SLA) while leaving the rest of the architecture largely intact. We then apply FP8-aware QAT with partial attention distillation to preserve geometric priors under low-cost computation~\cite{zhang2025sla,micikevicius2022fp8,hinton2015distilling}.

\begin{figure}[t]
  \centering
  \includegraphics[width=\linewidth]{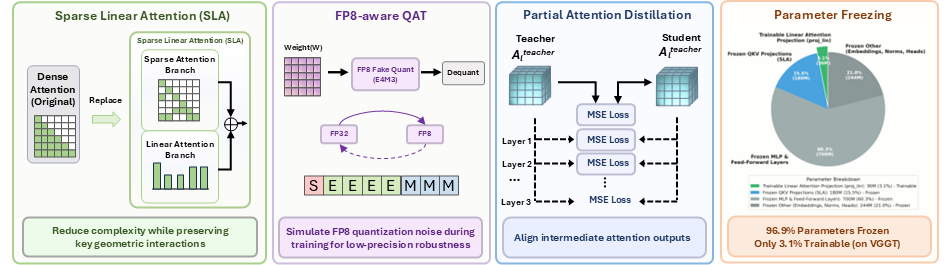}
  \caption{\textbf{Overall framework of Lite3R.} Starting from a dense pretrained 3D reconstruction teacher, Lite3R constructs a lite student by replacing dense attention with Sparse Linear Attention, freezing the inherited backbone projections, and training only lightweight linear-branch projection layers under FP8-aware quantization-aware training. Partial attention distillation preserves intermediate geometric priors, and the resulting student is converted to an efficient FP8-compatible deployment model.}
  \label{fig:main-framework}
\end{figure}

The pipeline is sequential: SLA reduces token-mixing cost, FP8-aware QAT enables stable low-precision deployment, and partial attention distillation aligns intermediate representations with the dense teacher. Figure~\ref{fig:main-framework} summarizes the framework and deployment pathway.

\subsection{Dense teacher and lite student construction}

We instantiate Lite3R on VGGT and DA3-Large, although the design is model-agnostic~\cite{wang2025vggt,lin2025depthanything3}. For each backbone, the dense pretrained model is the frozen teacher. The lite student copies teacher weights and replaces standard or memory-efficient attention with SLA blocks, while preserving geometry-critical components such as normalization, positional encoding, and task heads whenever possible~\cite{su2021roformer,ba2016layernorm}. We freeze most inherited backbone parameters and optimize mainly the lightweight linear-branch projection layers together with the quantization-aware linear path, reducing drift from the teacher feature space and stabilizing low-precision adaptation.

\subsection{Sparse Linear Attention for geometry backbones}

SLA serves as the structural lightweighting module in Lite3R. Since it is not our main novelty, we summarize only the system-relevant design here and defer a compact algorithm summary to Appendix~\ref{app:sla-summary}. Given input tokens $X \in \mathbb{R}^{N \times d}$ with projections $Q=XW_Q$, $K=XW_K$, and $V=XW_V$, standard self-attention computes $A_{\mathrm{full}}(Q,K,V)=\operatorname{Softmax}\!\left(\frac{QK^{\top}}{\sqrt{d}}\right)V,$ which is expensive for long multi-view token sequences. Lite3R replaces it with an SLA module of the form
\begin{equation}
A_{\mathrm{SLA}}(Q,K,V)=A_{\mathrm{sparse}}(Q,K,V)+\operatorname{Proj}\big(A_{\mathrm{lin}}(Q,K,V)\big),
\label{eq:sla-main}
\end{equation}
where a sparse branch preserves high-value geometric correspondences and a linear branch supplies low-cost global context. This replacement lowers token-mixing cost while maintaining a reasonable approximation to dense multi-view interaction~\cite{katharopoulos2020transformers,zhang2025sla}. SLA therefore defines the lightweight student architecture for FP8-aware adaptation.

\subsection{FP8-aware quantization-aware training}

The main methodological question in Lite3R is how to make a geometry-sensitive 3D reconstruction model robust under low precision. Replacing dense attention alone is insufficient because large linear layers and their activations still dominate memory traffic, and naive low-precision conversion can destabilize depth, pose, and 3D consistency. Lite3R therefore performs FP8-aware quantization-aware training (FP8-aware QAT) on the lite student. We use the E4M3 FP8 format throughout training and deployment, and inject FP8 perturbations during training so the student learns to operate under low-precision weight and activation noise; additional details are provided in Appendix~\ref{appendix:fp8_qat}.

This design matters because geometry errors can accumulate across long feature streams, the student already differs structurally from the teacher after SLA replacement, and our goal is to preserve pretrained geometric priors while translating them into a deployment-oriented computation path~\cite{micikevicius2022fp8,jacob2018quantization}. FP8-aware QAT is therefore the core adaptation mechanism in Lite3R.

\paragraph{Selective parameter freezing.}
FP8-aware QAT in Lite3R follows a parameter-efficient adaptation strategy. During training, only the lightweight linear-branch projection layers introduced by SLA are updated, while all original pretrained backbone parameters---including the \texttt{qkv} projections, MLP blocks, and other linear projections---remain frozen. For VGGT, only about 36M of 1.16B parameters ($\approx 3.1\%$) are trainable. We treat freezing primarily as a systems design choice: it reduces optimizer state and activation-related training memory, lowers update cost, and makes adaptation easier to scale across large backbones and longer token sequences.

All linear layers in the student, including frozen backbone layers, still participate in FP8 fake quantization during the forward pass so that the full computation graph experiences realistic low-precision perturbations. In the backward pass, gradients are applied only to the linear-branch projection layers, which keeps optimization lightweight and improves throughput while preserving compatibility with parameter-efficient adaptation recipes~\cite{hu2021lora}.

\paragraph{FP8 fake quantization of linear layers.}
During training, the linear layers in the student are replaced with FP8 fake-quantized versions. Let $W$ and $X$ denote the higher-precision weight and input activation (e.g., FP16/BF16). The forward pass simulates FP8 E4M3 quantization as
\begin{equation}
W_q=Q_{\mathrm{fp8}}(W), \qquad X_q=Q_{\mathrm{fp8}}(X), \qquad Y=\operatorname{Linear}(X_q,W_q),
\label{eq:fp8-fq}
\end{equation}
where $Q_{\mathrm{fp8}}(\cdot)$ denotes fake quantization with FP8 casting and dequantization in the forward path. In our implementation, weight quantization uses per-output-row dynamic scaling, activation quantization uses per-token dynamic scaling, and the backward pass adopts a straight-through estimator~\cite{jacob2018quantization,bengio2013stochastic}.

\paragraph{Mixed-precision treatment for geometry-sensitive operators.}
FP8-aware QAT does not force every operator into low precision. Geometry-sensitive components such as LayerNorm, positional encoding, RoPE, and selected non-linear operators remain in higher precision when needed. This mixed treatment preserves numerically fragile geometric computations while still pushing the dominant linear path toward an FP8-compatible regime~\cite{su2021roformer,ba2016layernorm,micikevicius2022fp8}.

\paragraph{Why FP8-aware QAT is needed in 3D reconstruction.}
3D reconstruction is more sensitive to quantization noise than many standard vision tasks. Small perturbations in intermediate features can propagate into multi-view matching, camera pose estimation, and point-cloud geometry. FP8-aware QAT mitigates this issue by exposing the student to realistic low-precision perturbations throughout optimization, allowing it to rebalance internal representations before deployment.

\subsection{Partial attention distillation and task supervision}

The student is trained with both the original geometry task objective and a partial attention distillation objective. The teacher is the frozen dense pretrained backbone, while the student is the SLA-based FP8-aware lite model. Rather than distilling final outputs such as depth, pose, or point clouds, Lite3R aligns intermediate attention-module outputs so that the student remains close to the teacher's internal geometric representation after structural replacement and quantization perturbation.

This design is tightly coupled with selective parameter freezing. Because FP8-aware QAT updates only lightweight linear-branch projection layers while keeping the original backbone frozen, training remains memory-efficient and scalable even for billion-parameter backbones. Partial attention distillation then guides the trainable layers to absorb the discrepancy caused by SLA replacement and FP8 perturbation while staying close to the teacher's intermediate responses~\cite{hinton2015distilling,zagoruyko2017at,hu2021lora}.

\paragraph{Partial attention distillation.}
For each selected attention-like module $l$, we register forward hooks on both teacher and student and record their output tensors $A_l^{\mathrm{teacher}}$ and $A_l^{\mathrm{student}}$. The distillation loss is defined as
\begin{equation}
\mathcal{L}_{\mathrm{attnKD}} = \frac{1}{L}\sum_{l=1}^{L}
\operatorname{MSE}\big(A_l^{\mathrm{student}},\operatorname{stopgrad}(A_l^{\mathrm{teacher}})\big),
\label{eq:attn-align}
\end{equation}
where $L$ is the number of aligned modules. This objective encourages the lite student to preserve the teacher's intermediate geometry-aware response patterns under both structural and numerical changes.

\paragraph{Joint training objective.}
Let $\mathcal{L}_{\mathrm{task}}$ denote the original geometry supervision used by the corresponding backbone. The overall training target is
\begin{equation}
\mathcal{L}_{\mathrm{total}} = \mathcal{L}_{\mathrm{task}} + \gamma\,\mathcal{L}_{\mathrm{attnKD}},
\label{eq:total-loss}
\end{equation}
where $\gamma$ is a fixed distillation coefficient. In the main Lite3R setting, we use a small constant weight to keep the student close to the dense teacher while allowing it to adapt to its own SLA and FP8-aware computation path. For DA3-Large and VGGT, $\mathcal{L}_{\mathrm{task}}$ follows the original geometry task definition of the corresponding backbone after adapting the output interface when necessary.

Task loss keeps final predictions aligned with dataset annotations, whereas attention distillation preserves the teacher's internal geometric representation.

\subsection{Deployment pathway}

After training, the FP8-aware student is converted into a deployment model by removing fake-quant modules and applying the available FP8 inference backend to the trained linear weights. Consistent with training, the deployed FP8 pathway also uses the E4M3 FP8 format. Under the current hardware runtime constraint, the stable path is FP8 \emph{weight-only} inference, even though training simulates both FP8 weight and activation perturbations. We therefore describe the method as \emph{FP8-aware QAT with an FP8 weight-only deployment backend}, which reflects the implemented system while preserving the main benefit of QAT: the student has already adapted during training to the low-precision regime expected at deployment.

Overall, Lite3R unifies SLA-based structural lightweighting, FP8-aware QAT, partial attention distillation, and an FP8-compatible deployment pathway in a model-agnostic framework that preserves the geometric strengths of modern 3D backbones while reducing inference and memory cost.

\section{Experiments}

We evaluate Lite3R on two representative geometry backbones, VGGT and DA3-Large, under a unified single-GPU setting. Our experiments answer four questions: (1) whether Lite3R preserves reconstruction quality after replacing dense attention with SLA, (2) whether the proposed FP8-aware route improves deployment efficiency in practice, (3) which components are most responsible for retaining geometry, and (4) how sensitive the method is to the distillation coefficient and fine-tuning schedule.

\subsection{Experimental setup}

\paragraph{Dataset and model.}
We use two datasets: BlendedMVS low-resolution and DTU64. BlendedMVS provides images, camera parameters, and depth supervision, so we report depth, pose, point-cloud geometry, and efficiency metrics on this benchmark~\cite{yao2020blendedmvs}. DTU64 is treated as a pose-oriented benchmark, so we report rotation/translation errors together with deployment efficiency~\cite{jensen2014dtu}. We evaluate two pretrained backbones, VGGT and Depth Anything 3 Large (DA3-Large), to test whether Lite3R generalizes across heterogeneous transformer architectures~\cite{wang2025vggt,lin2025depthanything3}.

\paragraph{Compared variants.}
Our main comparison is between the original backbone and \textbf{Lite3R}, which combines SLA with sparse-attention sparsity 0.2, FP8-aware QAT, and partial attention distillation, and is deployed with an E4M3 FP8 weight-only path~\cite{micikevicius2022fp8}. In the ablation study, we additionally compare SLA without FP8-aware QAT, no-SLA variants, and different distillation coefficients. Following our lightweight adaptation protocol, we freeze the original pretrained backbone and optimize only the lightweight linear-branch projection layers inside SLA, in the spirit of parameter-efficient adaptation~\cite{hu2021lora}.

\paragraph{Hardware settings.}
All experiments are conducted on a single \textbf{NVIDIA A100-PCIE-40GB}. We measure efficiency under the same evaluation scripts and input settings for the baseline and Lite3R. Although we discuss consumer-GPU deployment implications, all quantitative results here are measured on A100 to keep the comparison controlled.

\paragraph{Metrics.}
We jointly evaluate quality and efficiency. For depth, we report AbsRel, $\delta_1$, and RMSE when ground-truth depth is available. For pose, we report rotation and translation errors. For geometry, we report Chamfer distance and F-score at 5cm, following common reconstruction benchmarks~\cite{knapitsch2017tanks}. For efficiency, we report end-to-end mean latency and peak GPU memory. This setup measures whether Lite3R maintains competitive reconstruction metrics while reducing practical deployment cost.

\subsection{Main Results}

Tables~\ref{tab:main-results-blended} and~\ref{tab:main-results-dtu} compare the original backbones and Lite3R on BlendedMVS and DTU64, respectively. The trend is consistent across all four backbone--dataset pairs: Lite3R substantially reduces latency and memory while keeping downstream geometry metrics within an acceptable range. Our deployed model should be understood as \emph{FP8-aware QAT plus FP8 weight-only inference}: under the current code path on A100, native dynamic FP8 activation inference is not used.

\begin{figure}[t]
  \centering
  \includegraphics[width=\linewidth]{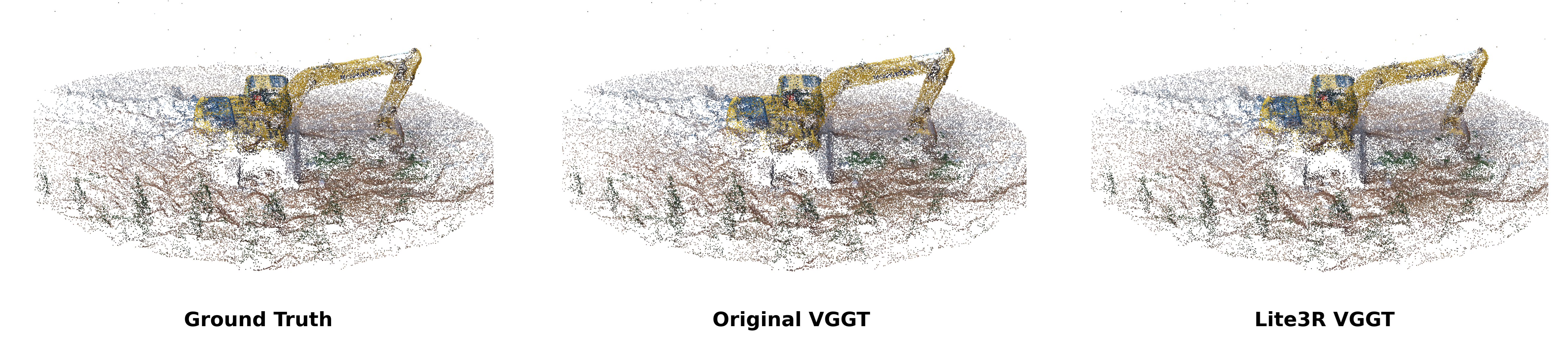}
  \caption{Qualitative point-cloud comparison on BlendedMVS between ground truth, the original VGGT backbone, and Lite3R instantiated on VGGT. Lite3R maintains the dominant scene layout and point-cloud structure while providing substantially better deployment efficiency than the dense teacher.}
  \label{fig:pointcloud-comparison}
\end{figure}

On BlendedMVS, VGGT-based Lite3R achieves 1.76$\times$ speedup and 2.32$\times$ memory saving, while AbsRel increases from 0.0184 to 0.0271 and rotation error increases from 1.9308 to 2.2300. DA3-Large-based Lite3R achieves even stronger efficiency gains (1.97$\times$ speedup, 1.98$\times$ memory saving) with comparable quality degradation. On DTU64, the same trend holds. The degradation mainly comes from two sources: SLA removes some long-range interactions for efficiency, and FP8 quantization introduces perturbations into geometry-sensitive computations. Even so, the degradation remains bounded and acceptable for deployment-oriented settings. Figure~\ref{fig:pointcloud-comparison} shows a qualitative point-cloud comparison on BlendedMVS, where Lite3R maintains the main scene geometry and global structure relative to the VGGT teacher and ground truth.

\begin{figure}[t]
  \centering
  \includegraphics[width=\linewidth]{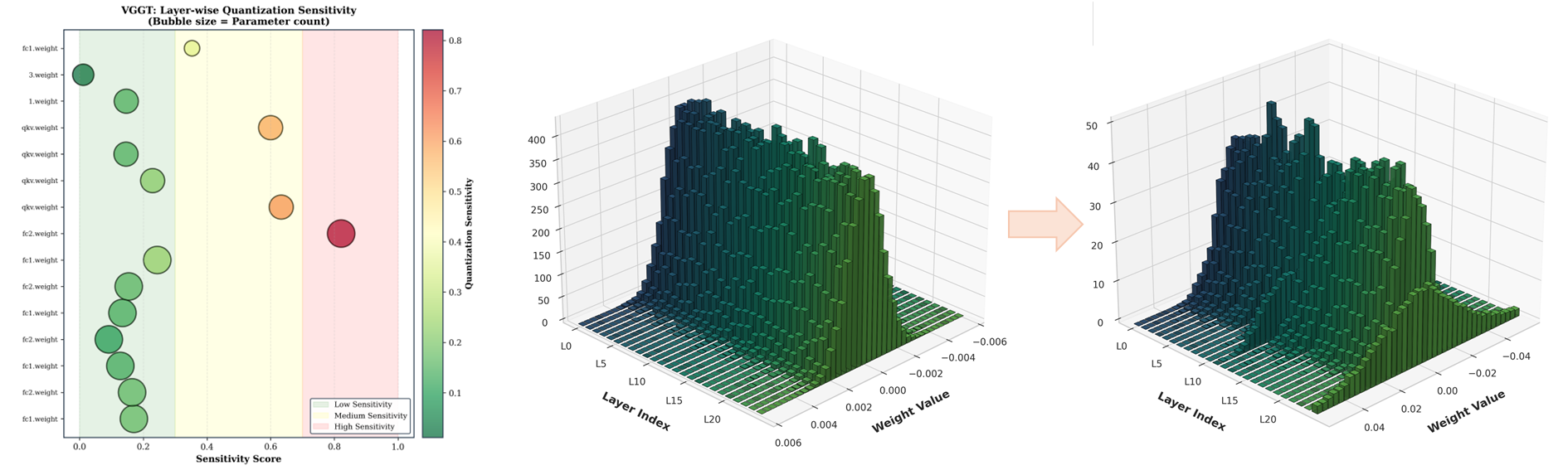}
  \caption{\textbf{Analysis of Lite3R adaptation sensitivity.} Left: layer-wise quantization sensitivity of VGGT, showing that different backbone stages respond unevenly to low-precision perturbations. Right: change pattern of the linear-branch projection layers during training, illustrating how continued FP8-aware QAT increases drift in the small trainable subspace and helps explain the weaker stability of longer schedules. Appendix~\ref{appendix:main-results-comprehensive} provides details on sensitivity scores.}
  \label{fig:proj-lin-change}
\end{figure}

\begin{figure}[t]
  \centering
  \includegraphics[width=\linewidth]{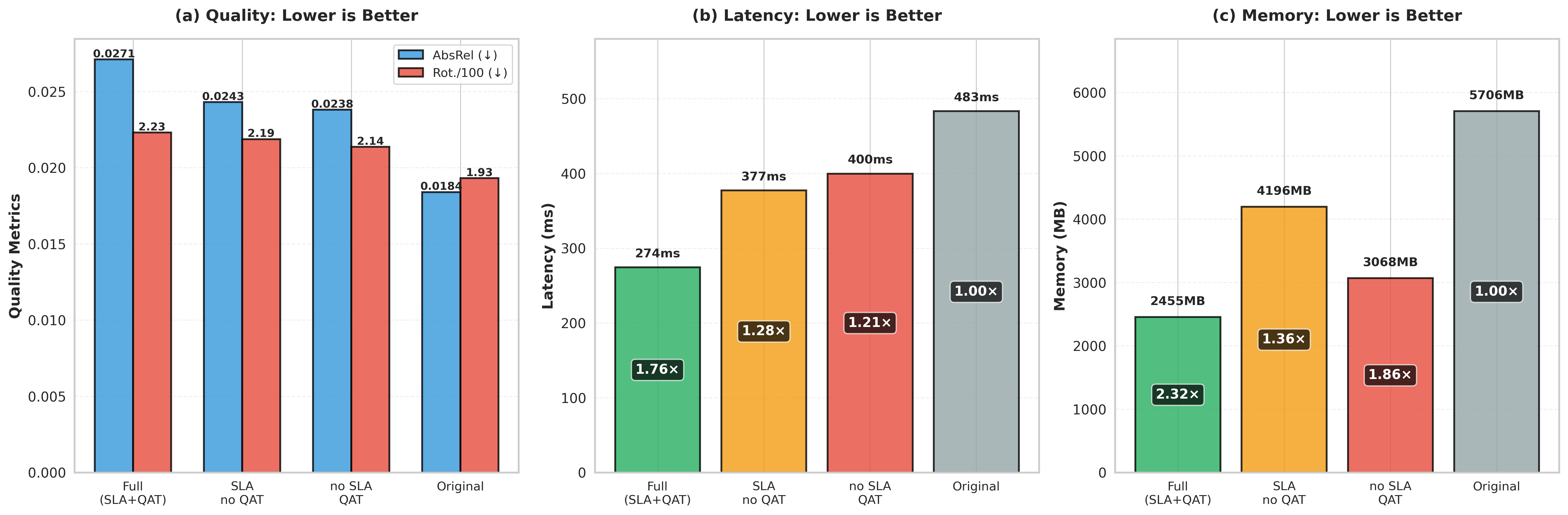}
  \caption{\textbf{Visual comparison of the component-ablation results on VGGT over BlendedMVS.} The chart summarizes Table~\ref{tab:component-ablation} and shows that removing SLA or FP8-aware QAT recovers only part of the final quality-efficiency tradeoff.}
  \label{fig:component-ablation}
\end{figure}

\begin{table}[t]
  \caption{Main results on BlendedMVS. Metric directions are indicated in the table header.}
  \label{tab:main-results-blended}
  \centering
  \scriptsize
  \setlength{\tabcolsep}{3pt}
  \makebox[\linewidth][c]{%
  \begin{tabular}{llcccccccccc}
    \toprule
    Method & Backbone & AbsRel$\downarrow$ & $\delta_1$$\uparrow$ & Rot.$\downarrow$ & Trans.$\downarrow$ & CD$\downarrow$ & F5cm$\uparrow$ & Latency$\downarrow$ & Memory$\downarrow$ & Speedup$\uparrow$ & Memory Saving$\uparrow$ \\
    \midrule
    Original & VGGT & 0.0184 & 0.9930 & 1.9308 & 0.0273 & 0.2411 & 0.2005 & 483.33 & 5706 & 1.00$\times$ & 1.00$\times$ \\
    Lite3R & VGGT & 0.0271 & 0.9922 & 2.2300 & 0.0285 & 0.2354 & 0.2029 & 274.38 & 2455 & 1.76$\times$ & 2.32$\times$ \\
    Original & DA3-Large & 0.0862 & 0.9329 & 9.4800 & 0.0838 & 0.5366 & 0.1149 & 187.29 & 2713 & 1.00$\times$ & 1.00$\times$ \\
    Lite3R & DA3-Large & 0.0889 & 0.9308 & 10.7440 & 0.1239 & 0.5892 & 0.1210 & 95.27 & 1368 & 1.97$\times$ & 1.98$\times$ \\
    \bottomrule
  \end{tabular}%
  }
\end{table}

\begin{table}[t]
  \caption{Main results on DTU64, which is currently treated as a pose-oriented benchmark. Metric directions are indicated in the table header.}
  \label{tab:main-results-dtu}
  \centering
  \scriptsize
  \setlength{\tabcolsep}{6pt}
  \makebox[\linewidth][c]{%
  \begin{tabular}{llcccccc}
    \toprule
    Backbone & Method & Rot.$\downarrow$ & Trans.$\downarrow$ & Latency$\downarrow$ & Memory$\downarrow$ & Speedup$\uparrow$ & Memory Saving$\uparrow$ \\
    \midrule
    VGGT & Original & 0.3811 & 0.0192 & 482.45 & 5701 & 1.00$\times$ & 1.00$\times$ \\
    VGGT & Lite3R & 0.7003 & 0.0220 & 275.98 & 2452 & 1.75$\times$ & 2.33$\times$ \\
    DA3-Large & Original & 0.9313 & 0.0119 & 186.65 & 2709 & 1.00$\times$ & 1.00$\times$ \\
    DA3-Large & Lite3R & 1.5786 & 0.0211 & 99.54 & 1364 & 1.87$\times$ & 1.99$\times$ \\
    \bottomrule
  \end{tabular}%
  }
\end{table}

VGGT shows the strongest quality--efficiency tradeoff. On BlendedMVS, Lite3R keeps $\delta_1$ nearly unchanged (0.9930 to 0.9922), slightly improves Chamfer distance (0.2411 to 0.2354) and F5cm (0.2005 to 0.2029), while reducing latency from 483.33ms to 274.38ms and memory from 5706MB to 2455MB. This corresponds to 1.76$\times$ speedup and 2.32$\times$ memory saving. On DTU64, the pose error increases from 0.3811/0.0192 to 0.7003/0.0220 in rotation/translation, but latency and memory still improve to 275.98ms and 2452MB, corresponding to 1.75$\times$ speedup and 2.33$\times$ memory saving.

DA3-Large is more sensitive than VGGT, but still benefits substantially from Lite3R. On BlendedMVS, Lite3R increases AbsRel slightly from 0.0862 to 0.0889, while $\delta_1$ remains close to the baseline (0.9329 to 0.9308) and F5cm improves from 0.1149 to 0.1210. Pose and Chamfer fluctuate more than on VGGT, indicating that DA3-Large is less robust to structural replacement. Even so, latency drops from 187.29ms to 95.27ms and memory from 2713MB to 1368MB, corresponding to 1.97$\times$ speedup and 1.98$\times$ memory saving. On DTU64, the same trend holds: pose error increases to 1.5786/0.0211, but latency and memory improve to 99.54ms and 1364MB, corresponding to 1.87$\times$ speedup and 1.99$\times$ memory saving. Appendix~\ref{appendix:main-results-comprehensive} provides additional context on the DA3-Large parameter allocation after SLA replacement, showing how its extremely small trainable subspace relates to this sharper quality--efficiency tradeoff.

\subsection{Ablation Study}

\paragraph{SLA and FP8-aware QAT}

We next examine which components are essential for the final behavior. Table~\ref{tab:component-ablation} compares the full recipe with two simplified variants, while Figure~\ref{fig:component-ablation} visualizes the same quality--efficiency tradeoff.

We restrict the component ablation to VGGT on BlendedMVS, where all geometry and efficiency metrics are directly comparable. Removing FP8-aware QAT improves quality slightly (AbsRel 0.0243 vs.0.0271; rotation 2.1866 vs.2.23030) but degrades efficiency sharply, increasing latency/memory from 274.38ms/2455MB to 377.21ms/4196MB. Removing SLA yields the opposite pattern: quality remains competitive (AbsRel 0.0238; rotation 2.1374), but speedup and memory saving drop to only 1.21$\times$ and 1.86$\times$. Overall, Figure~\ref{fig:component-ablation} makes the trend clear: \textit{SLA mainly drives latency reduction, FP8-aware QAT mainly drives memory efficiency, and the full Lite3R recipe gives the best overall tradeoff}.

\paragraph{Distillation coefficient ablation}

We further vary the distillation coefficient over $\gamma\in\{0, 0.1, 0.2, 0.5\}$ to test how strongly the student should track the teacher.

\begin{table}[t]
  \caption{Component ablation on VGGT over BlendedMVS. Full denotes the main Lite3R recipe with SLA, FP8-aware QAT, and KD coefficient 0.1. Metric directions are indicated in the table header.}
  \label{tab:component-ablation}
  \centering
  \scriptsize
  \setlength{\tabcolsep}{3pt}
  \makebox[\linewidth][c]{%
  \begin{tabular}{lcccccccccc}
    \toprule
    Variant & AbsRel$\downarrow$ & $\delta_1$$\uparrow$ & Rot.$\downarrow$ & Trans.$\downarrow$ & CD$\downarrow$ & F5cm$\uparrow$ & Latency$\downarrow$ & Memory$\downarrow$ & Speedup$\uparrow$ & Memory Saving$\uparrow$ \\
    \midrule
    SLA, QAT & 0.0271 & 0.9922 & 2.2300 & 0.0285 & 0.2354 & 0.2029 & 274.38 & 2455 & 1.76$\times$ & 2.32$\times$ \\
    SLA, no QAT & 0.0243 & 0.9924 & 2.1866 & 0.0274 & 0.2391 & 0.2042 & 377.21 & 4196 & 1.28$\times$ & 1.36$\times$ \\
    no SLA, QAT & 0.0238 & 0.9925 & 2.1374 & 0.0279 & 0.2397 & 0.2001 & 399.76 & 3068 & 1.21$\times$ & 1.86$\times$ \\
    Original & 0.0184 & 0.9930 & 1.9308 & 0.0273 & 0.2411 & 0.2005 & 483.33 & 5706 & 1.00$\times$ & 1.00$\times$ \\
    \bottomrule
  \end{tabular}%
  }
\end{table}

\begin{table}[t!]
  \caption{Distillation-coefficient ablation on VGGT over BlendedMVS for SLA+FP8-aware QAT. The $\gamma=0.1$ row corresponds to the main Lite3R setting in Table~\ref{tab:main-results-blended}. Metric directions are indicated in the table header.}
  \label{tab:kd-ablation}
  \centering
  \scriptsize
  \setlength{\tabcolsep}{4pt}
  \makebox[\linewidth][c]{%
  \begin{tabular}{ccccccccccc}
    \toprule
    $\gamma$ & AbsRel$\downarrow$ & $\delta_1$$\uparrow$ & Rot.$\downarrow$ & Trans.$\downarrow$ & CD$\downarrow$ & F5cm$\uparrow$ & Latency$\downarrow$ & Memory$\downarrow$ & Speedup$\uparrow$ & Memory Saving$\uparrow$ \\
    \midrule
    0.0 & 0.0283 & 0.9926 & 2.2817 & 0.0297 & 0.2305 & 0.2003 & 276.22 & 2457 & 1.75$\times$ & 2.32$\times$ \\
    0.1 & 0.0271 & 0.9922 & 2.2300 & 0.0285 & 0.2354 & 0.2029 & 274.38 & 2455 & 1.76$\times$ & 2.32$\times$ \\
    0.2 & 0.0291 & 0.9922 & 2.2663 & 0.0288 & 0.2247 & 0.1938 & 274.90 & 2457 & 1.76$\times$ & 2.32$\times$ \\
    0.5 & 0.0272 & 0.9924 & 2.2640 & 0.0292 & 0.2238 & 0.1900 & 272.56 & 2457 & 1.77$\times$ & 2.32$\times$ \\
    \bottomrule
  \end{tabular}%
  }
\end{table}

We again restrict the study to VGGT on BlendedMVS. The trend is non-monotonic, but $\gamma=0.1$ is the most balanced choice: it gives the best AbsRel, rotation error, translation error, and F5cm, while all settings remain nearly identical in efficiency (about 1.75$\times$--1.77$\times$ speedup and 2.32$\times$ memory saving). This supports using a small KD coefficient as a lightweight auxiliary constraint rather than a dominant supervision term.

\paragraph{Training schedule analysis}

We also study a longer 20-epoch FP8-aware QAT schedule under the same frozen-backbone, linear-projection-layer-only setting. Although training completes on both VGGT and DA3-Large, the resulting checkpoints show clear geometric drift, especially in pose, Chamfer distance, and F5cm. Figure~\ref{fig:proj-lin-change} suggests that prolonged optimization moves the lightweight adaptation layers away from the most stable regime. We therefore use the shorter 1-epoch FP8-aware QAT checkpoint in the main results.

\paragraph{Deployment discussion}

The results above support two deployment-level conclusions. First, Lite3R consistently reduces the end-to-end memory footprint by about $1.9$ to $2.4\times$ across both backbone families, which is critical for long multi-view inputs. Second, lower numerical precision alone does not determine practical speed; end-to-end latency also depends on kernel availability, graph compilation, dequantization overhead, tensor-core support, and memory movement. Our reported speedups are measured on an NVIDIA A100, where the current runtime path does not exploit hardware-specialized FP8 inference as aggressively as newer deployment GPUs. On GPUs with stronger FP8-oriented support, such as H20-class accelerators, the same Lite3R deployment path should have additional headroom for speedup. This is why we frame our method as an algorithm--system co-design: SLA reduces token-mixing cost, while FP8-aware QAT and weight-only deployment translate that reduction into a stable, measurable efficiency gain.

\section{Conclusion}

We presented \textbf{Lite3R}, a model-agnostic framework for efficient feed-forward 3D reconstruction that combines Sparse Linear Attention, parameter-efficient FP8-aware QAT, and partial attention distillation. By replacing dense attention and adapting only lightweight linear-branch projection layers, Lite3R converts dense pretrained geometry backbones into deployment-oriented low-precision models. Experiments on VGGT and DA3-Large over BlendedMVS and DTU64 show that Lite3R reduces latency (1.76--1.97$\times$) and memory footprint (2.32--2.71$\times$) while maintaining competitive depth, pose, and 3D reconstruction quality overall; ablations further show that both SLA and FP8-aware QAT are important for the best quality--efficiency tradeoff. Overall, Lite3R provides a practical algorithm--system co-design approach to scalable transformer-based 3D reconstruction under realistic hardware constraints.

\clearpage
\bibliographystyle{plain}
\bibliography{main}

@inproceedings{wang2025vggt,
  title = {VGGT: Visual Geometry Grounded Transformer},
  author = {Wang, Jianyuan and Chen, Minghao and Karaev, Nikita and Vedaldi, Andrea and Rupprecht, Christian and Novotny, David},
  booktitle = {Computer Vision and Pattern Recognition},
  year = {2025},
  journal = {Computer Vision and Pattern Recognition},
  pages = {5294-5306},
  doi = {10.1109/CVPR52734.2025.00499},
  publisher = {IEEE},
}

@article{lin2025depthanything3,
  title = {Depth Anything 3: Recovering the Visual Space from Any Views},
  author = {Lin, Haotong and Chen, Sili and Liew, Jun Hao and Chen, Donny Y. and Li, Zhenyu and Shi, Guang and Feng, Jiashi and Kang, Bingyi},
  journal = {arXiv.org},
  year = {2025},
  doi = {10.48550/arXiv.2511.10647},
}

@article{wang2024dust3r,
  title = {DUSt3R: Geometric 3D Vision Made Easy},
  author = {Wang, Shuzhe and Leroy, Vincent and Cabon, Yohann and Chidlovskii, Boris and Revaud, Jerome},
  journal = {Computer Vision and Pattern Recognition},
  year = {2023},
  doi = {10.1109/CVPR52733.2024.01956},
}

@inproceedings{leroy2024mast3r,
  title = {Grounding Image Matching in 3D with MASt3R},
  author = {Leroy, Vincent and Cabon, Yohann and Revaud, Jerome},
  booktitle = {European Conference on Computer Vision},
  year = {2024},
  journal = {European Conference on Computer Vision},
  doi = {10.48550/arXiv.2406.09756},
}

@article{wang2023vggsfm,
  title = {VGGSfM: Visual Geometry Grounded Deep Structure From Motion},
  author = {Wang, Jianyuan and Karaev, Nikita and Rupprecht, Christian and Novotny, David},
  journal = {Computer Vision and Pattern Recognition},
  year = {2023},
  doi = {10.1109/CVPR52733.2024.02049},
}

@article{zhang2025sla,
  title = {SLA: Beyond Sparsity in Diffusion Transformers via Fine-Tunable Sparse-Linear Attention},
  author = {Zhang, Jintao and Wang, Haoxu and Jiang, Kai and Yang, Shuo and Zheng, Kaiwen and Xi, Haocheng and Wang, Ziteng and Zhu, Hongzhou and Zhao, Min and Stoica, Ion and others},
  journal = {arXiv.org},
  year = {2025},
  doi = {10.48550/arXiv.2509.24006},
}

@inproceedings{katharopoulos2020transformers,
  title = {Transformers are RNNs: Fast Autoregressive Transformers with Linear Attention},
  author = {Katharopoulos, Angelos and Vyas, Apoorv and Pappas, Nikolaos and Fleuret, Fran{\c{c}}ois},
  booktitle = {International Conference on Machine Learning},
  pages = {5156--5165},
  year = {2020},
  journal = {International Conference on Machine Learning},
}

@article{dao2022flashattention,
  title = {FlashAttention: Fast and Memory-Efficient Exact Attention with IO-Awareness},
  author = {Dao, Tri and Fu, Daniel Y. and Ermon, Stefano and Rudra, A. and R'e, Christopher},
  journal = {Neural Information Processing Systems},
  year = {2022},
  pages = {16344-16359},
  doi = {10.52202/068431-1189},
  publisher = {Neural Information Processing Systems Foundation, Inc. (NeurIPS)},
}

@inproceedings{shah2024flashattention3,
  title = {FlashAttention-3: Fast and Accurate Attention with Asynchrony and Low-precision},
  author = {Shah, Jay and Bikshandi, Ganesh and Zhang, Ying and Thakkar, Vijay and Ramani, Pradeep and Dao, Tri},
  booktitle = {Neural Information Processing Systems},
  volume = {37},
  year = {2024},
  journal = {Neural Information Processing Systems},
  pages = {68658-68685},
  doi = {10.48550/arXiv.2407.08608},
  publisher = {Neural Information Processing Systems Foundation, Inc. (NeurIPS)},
}

@inproceedings{vaswani2017attention,
  title = {Attention is All You Need},
  author = {Vaswani, Ashish and Shazeer, Noam and Parmar, Niki and Uszkoreit, Jakob and Jones, Llion and Gomez, Aidan N and Kaiser, {\L}ukasz and Polosukhin, Illia},
  booktitle = {Neural Information Processing Systems},
  volume = {30},
  pages = {5998--6008},
  year = {2017},
  journal = {Neural Information Processing Systems},
  doi = {10.65215/nxvz2v36},
  publisher = {Shenzhen Medical Academy of Research and Translation},
}

@article{dosovitskiy2020vit,
  title = {An Image is Worth 16x16 Words: Transformers for Image Recognition at Scale},
  author = {Dosovitskiy, Alexey and Beyer, Lucas and Kolesnikov, Alexander and Weissenborn, Dirk and Zhai, Xiaohua and Unterthiner, Thomas and Dehghani, Mostafa and Minderer, Matthias and Heigold, Georg and Gelly, Sylvain and others},
  journal = {International Conference on Learning Representations},
  year = {2020},
}

@article{oquab2023dinov2,
  title = {DINOv2: Learning Robust Visual Features without Supervision},
  author = {Oquab, M. and Darcet, Timothée and Moutakanni, Théo and Vo, Huy V. and Szafraniec, Marc and Khalidov, Vasil and Fernandez, Pierre and Haziza, Daniel and Massa, Francisco and El-Nouby, Alaaeldin and others},
  journal = {Trans. Mach. Learn. Res.},
  year = {2023},
  doi = {10.48550/arXiv.2304.07193},
}

@article{micikevicius2022fp8,
  title = {FP8 Formats for Deep Learning},
  author = {Micikevicius, P. and Stosic, Dusan and Burgess, N. and Cornea, Marius and Dubey, P. and Grisenthwaite, R. and Ha, Sangwon and Heinecke, A. and Judd, Patrick and Kamalu, John and others},
  journal = {arXiv.org},
  year = {2022},
  doi = {10.48550/arXiv.2209.05433},
}

@inproceedings{jacob2018quantization,
  title = {Quantization and Training of Neural Networks for Efficient Integer-Arithmetic-Only Inference},
  author = {Jacob, Benoit and Kligys, Skirmantas and Chen, Bo and Zhu, Menglong and Tang, Matthew and Howard, Andrew and Adam, Hartwig and Kalenichenko, Dmitry},
  booktitle = {2018 IEEE/CVF Conference on Computer Vision and Pattern Recognition},
  pages = {2704--2713},
  year = {2017},
  journal = {2018 IEEE/CVF Conference on Computer Vision and Pattern Recognition},
  doi = {10.1109/CVPR.2018.00286},
}

@article{hinton2015distilling,
  title = {Distilling the Knowledge in a Neural Network},
  author = {Hinton, Geoffrey and Vinyals, Oriol and Dean, Jeff},
  journal = {arXiv.org},
  year = {2015},
}

@inproceedings{zagoruyko2017at,
  title = {Paying More Attention to Attention: Improving the Performance of Convolutional Neural Networks via Attention Transfer},
  author = {Zagoruyko, Sergey and Komodakis, Nikos},
  booktitle = {International Conference on Learning Representations},
  year = {2016},
  journal = {International Conference on Learning Representations},
}

@inproceedings{yao2020blendedmvs,
  title = {BlendedMVS: A Large-Scale Dataset for Generalized Multi-View Stereo Networks},
  author = {Yao, Yao and Luo, Zixin and Li, Shiwei and Zhang, Jingyang and Ren, Yufan and Zhou, Lei and Fang, Tian and Quan, Long},
  booktitle = {Computer Vision and Pattern Recognition},
  pages = {1790--1799},
  year = {2019},
  journal = {Computer Vision and Pattern Recognition},
  doi = {10.1109/cvpr42600.2020.00186},
}

@inproceedings{jensen2014dtu,
  title = {Large Scale Multi-view Stereopsis Evaluation},
  author = {Jensen, R. and Dahl, A. and Vogiatzis, George and Tola, Engil and Aanæs, H.},
  booktitle = {2014 IEEE Conference on Computer Vision and Pattern Recognition},
  year = {2014},
  journal = {2014 IEEE Conference on Computer Vision and Pattern Recognition},
  pages = {406-413},
  doi = {10.1109/CVPR.2014.59},
  publisher = {IEEE},
}

@article{knapitsch2017tanks,
  title = {Tanks and Temples: Benchmarking Large-Scale Scene Reconstruction},
  author = {Knapitsch, Arno and Park, Jaesik and Zhou, Qian-Yi and Koltun, Vladlen},
  journal = {ACM Transactions on Graphics},
  volume = {36},
  number = {4},
  pages = {1--13},
  year = {2017},
}

@inproceedings{schonberger2016colmap,
  title = {Structure-from-Motion Revisited},
  author = {Schönberger, Johannes L. and Frahm, Jan-Michael},
  booktitle = {Computer Vision and Pattern Recognition},
  pages = {4104--4113},
  year = {2016},
  journal = {Computer Vision and Pattern Recognition},
  doi = {10.1109/CVPR.2016.445},
  publisher = {IEEE},
}

@inproceedings{pan2024glomap,
  title = {Global Structure-from-Motion Revisited},
  author = {Pan, Linfei and Baráth, Dániel and Pollefeys, M. and Schönberger, Johannes L.},
  booktitle = {European Conference on Computer Vision},
  year = {2024},
  journal = {European Conference on Computer Vision},
  doi = {10.48550/arXiv.2407.20219},
}

@inproceedings{yao2018mvsnet,
  title = {MVSNet: Depth Inference for Unstructured Multi-view Stereo},
  author = {Yao, Yao and Luo, Zixin and Li, Shiwei and Fang, Tian and Quan, Long},
  booktitle = {European Conference on Computer Vision},
  year = {2018},
  journal = {European Conference on Computer Vision},
  pages = {785-801},
  doi = {10.1007/978-3-030-01237-3_47},
  publisher = {Springer International Publishing},
}

@inproceedings{chen2019pointmvsnet,
  title = {Point-Based Multi-View Stereo Network},
  author = {Chen, Rui and Han, Songfang and Xu, Jing and Su, Hao},
  booktitle = {IEEE International Conference on Computer Vision},
  year = {2019},
  journal = {IEEE International Conference on Computer Vision},
  pages = {1538-1547},
  doi = {10.1109/ICCV.2019.00162},
  publisher = {IEEE},
}

@inproceedings{vats2024gcmvsnet,
  title = {GC-MVSNet: Multi-View, Multi-Scale, Geometrically-Consistent Multi-View Stereo},
  author = {Vats, Vibhas K. and Joshi, Sripad and Crandall, David J. and Reza, Md. Alimoor and Jung, Soon-heung},
  booktitle = {IEEE Workshop/Winter Conference on Applications of Computer Vision},
  pages = {3242--3252},
  year = {2023},
  journal = {IEEE Workshop/Winter Conference on Applications of Computer Vision},
  doi = {10.1109/WACV57701.2024.00321},
}

@inproceedings{zhang2023ramvsnet,
  title = {Multi-View Stereo Representation Revist: Region-Aware MVSNet},
  author = {Zhang, Yisu and Zhu, Jianke and Lin, Lixiang},
  booktitle = {Computer Vision and Pattern Recognition},
  year = {2023},
  journal = {Computer Vision and Pattern Recognition},
  pages = {17376-17385},
  doi = {10.1109/CVPR52729.2023.01667},
  publisher = {IEEE},
}

@inproceedings{liao2022wtmvsnet,
  title = {WT-MVSNet: Window-based Transformers for Multi-view Stereo},
  author = {Liao, Jinli and Ding, Yikang and Shavit, Yoli and Huang, Dihe and Ren, Shihao and Guo, Jia and Feng, Wensen and Zhang, Kai},
  booktitle = {Neural Information Processing Systems},
  volume = {35},
  year = {2022},
  journal = {Neural Information Processing Systems},
  pages = {8564-8576},
  doi = {10.48550/arXiv.2205.14319},
  publisher = {Neural Information Processing Systems Foundation, Inc. (NeurIPS)},
}

@article{chen2024mvsplat,
  title = {MVSplat: Efficient 3D Gaussian Splatting from Sparse Multi-View Images},
  author = {Chen, Yuedong and Xu, Haofei and Zheng, Chuanxia and Zhuang, Bohan and Pollefeys, Marc and Geiger, Andreas and Cham, Tat-Jen and Cai, Jianfei},
  journal = {European Conference on Computer Vision},
  year = {2024},
  doi = {10.1007/978-3-031-72664-4_21},
}

@article{su2021roformer,
  title = {RoFormer: Enhanced Transformer with Rotary Position Embedding},
  author = {Su, Jianlin and Lu, Yu and Pan, Shengfeng and Wen, Bo and Liu, Yunfeng},
  journal = {Neurocomputing},
  year = {2021},
  doi = {10.1016/j.neucom.2023.127063},
}

@article{ba2016layernorm,
  title = {Layer Normalization},
  author = {Ba, Jimmy Lei and Kiros, Jamie Ryan and Hinton, Geoffrey E},
  journal = {arXiv.org},
  year = {2016},
}

@inproceedings{hu2021lora,
  title = {LoRA: Low-Rank Adaptation of Large Language Models},
  author = {Hu, Edward J and Shen, Yelong and Wallis, Phillip and Allen-Zhu, Zeyuan and Li, Yuanzhi and Wang, Shean and Wang, Lu and Chen, Weizhu},
  booktitle = {International Conference on Learning Representations},
  year = {2021},
  journal = {International Conference on Learning Representations},
}

@article{bengio2013stochastic,
  title = {Estimating or Propagating Gradients Through Stochastic Neurons for Conditional Computation},
  author = {Bengio, Yoshua and Léonard, Nicholas and Courville, Aaron C.},
  journal = {arXiv.org},
  year = {2013},
}

@inproceedings{yang2025fast3r,
  title = {Fast3R: Towards 3D Reconstruction of 1000+ Images in One Forward Pass},
  author = {Yang, Jianing and Sax, Alexander and Liang, Kevin J. and Henaff, Mikael and Tang, Hao and Cao, Ang and Chai, Joyce and Meier, Franziska and Feiszli, Matt},
  booktitle = {Computer Vision and Pattern Recognition},
  year = {2025},
  journal = {Computer Vision and Pattern Recognition},
  pages = {21924-21935},
  doi = {10.1109/CVPR52734.2025.02042},
  publisher = {IEEE},
}

@inproceedings{lan2026stream3r,
  title = {{STream3R}: Scalable Sequential {3D} Reconstruction with Causal Transformer},
  author = {Lan, Yushi and Luo, Yihang and Hong, Fangzhou and Zhou, Shangchen and Chen, Honghua and Lyu, Zhaoyang and Yang, Shuai and Dai, Bo and Loy, Chen Change and Pan, Xingang},
  booktitle = {arXiv.org},
  year = {2025},
  journal = {arXiv.org},
  doi = {10.48550/arXiv.2508.10893},
}

@inproceedings{xiao2023smoothquant,
  title = {SmoothQuant: Accurate and Efficient Post-Training Quantization for Large Language Models},
  author = {Xiao, Guangxuan and Lin, Ji and Seznec, Mickael and Demouth, Julien and Han, Song},
  booktitle = {International Conference on Machine Learning},
  year = {2022},
  journal = {International Conference on Machine Learning},
  doi = {10.48550/arXiv.2211.10438},
}

@inproceedings{lin2024awq,
  title = {AWQ: Activation-aware Weight Quantization for LLM Compression and Acceleration},
  author = {Lin, Ji and Tang, Jiaming and Tang, Haotian and Yang, Shang and Dang, Xingyu and Han, Song},
  booktitle = {arXiv.org},
  year = {2023},
  journal = {arXiv.org},
  doi = {10.48550/arXiv.2306.00978},
}

@inproceedings{frantar2023gptq,
  title = {GPTQ: Accurate Post-Training Quantization for Generative Pre-trained Transformers},
  author = {Frantar, Elias and Ashkboos, Saleh and Hoefler, Torsten and Alistarh, Dan},
  booktitle = {arXiv.org},
  year = {2022},
  journal = {arXiv.org},
}

@inproceedings{choromanski2021performer,
  title = {Rethinking Attention with Performers},
  author = {Choromanski, Krzysztof Marcin and Likhosherstov, Valerii and Dohan, David and Song, Xingyou and Gane, Andreea and Sarlos, Tamas and Hawkins, Peter and Davis, Jared Quincy and Mohiuddin, Afroz and Kaiser, Lukasz and others},
  booktitle = {International Conference on Learning Representations},
  year = {2020},
  journal = {International Conference on Learning Representations},
}

@article{wang2020linformer,
  title = {Linformer: Self-Attention with Linear Complexity},
  author = {Wang, Sinong and Li, Belinda and Khabsa, Madian and Fang, Han and Ma, Hao},
  journal = {arXiv.org},
  year = {2020},
}

@inproceedings{ranftl2021dpt,
  title = {Vision Transformers for Dense Prediction},
  author = {Ranftl, Ren\'{e} and Bochkovskiy, Alexey and Koltun, Vladlen},
  booktitle = {IEEE International Conference on Computer Vision},
  year = {2021},
  journal = {IEEE International Conference on Computer Vision},
  pages = {12159-12168},
  doi = {10.1109/ICCV48922.2021.01196},
  publisher = {IEEE},
}

@inproceedings{wu2022tinyvit,
  title = {TinyViT: Fast Pretraining Distillation for Small Vision Transformers},
  author = {Wu, Kan and Zhang, Jinnian and Peng, Houwen and Liu, Mengchen and Xiao, Bin and Fu, Jianlong and Yuan, Lu},
  booktitle = {European Conference on Computer Vision},
  year = {2022},
  journal = {European Conference on Computer Vision},
  pages = {68-85},
  doi = {10.48550/arXiv.2207.10666},
  publisher = {Springer Nature Switzerland},
}

@inproceedings{mehta2022mobilevit,
  title = {MobileViT: Light-weight, General-purpose, and Mobile-friendly Vision Transformer},
  author = {Mehta, Sachin and Rastegari, Mohammad},
  booktitle = {International Conference on Learning Representations},
  year = {2021},
  journal = {International Conference on Learning Representations},
}

@inproceedings{touvron2021deit,
  title = {Training data-efficient image transformers \& distillation through attention},
  author = {Touvron, Hugo and Cord, M. and Douze, Matthijs and Massa, Francisco and Sablayrolles, Alexandre and J'egou, Herv'e},
  booktitle = {International Conference on Machine Learning},
  year = {2020},
  journal = {International Conference on Machine Learning},
}

@inproceedings{murai2025mast3rslam,
  title = {MASt3R-SLAM: Real-Time Dense SLAM with 3D Reconstruction Priors},
  author = {Murai, Riku and Dexheimer, Eric and Davison, Andrew J.},
  booktitle = {Computer Vision and Pattern Recognition},
  year = {2024},
  journal = {Computer Vision and Pattern Recognition},
  doi = {10.1109/CVPR52734.2025.01556},
}

@inproceedings{duisterhof2025mast3rsfm,
  title = {MASt3R-SfM: a Fully-Integrated Solution for Unconstrained Structure-from-Motion},
  author = {Duisterhof, Bardienus Pieter and Zust, Lojze and Weinzaepfel, Philippe and Leroy, Vincent and Cabon, Yohann and Revaud, Jerome},
  booktitle = {International Conference on 3D Vision},
  year = {2024},
  journal = {International Conference on 3D Vision},
  doi = {10.1109/3DV66043.2025.00008},
}

@article{tang2024mv,
  title = {MV-DUSt3R+: Single-Stage Scene Reconstruction from Sparse Views In 2 Seconds},
  author = {Tang, Zhenggang and Fan, Yuchen and Wang, Dilin and Xu, Hongyu and Ranjan, Rakesh and Schwing, Alexander and Yan, Zhicheng},
  journal = {Computer Vision and Pattern Recognition},
  year = {2024},
  doi = {10.1109/CVPR52734.2025.00498},
}

@inproceedings{test3r2025,
  title = {Test3R: Test-Time Learning for Geometric 3D Vision},
  author = {Anonymous},
  booktitle = {Advances in Neural Information Processing Systems (NeurIPS)},
  year = {2025},
}

@inproceedings{vora2023divinet,
  title = {DiViNeT: 3D Reconstruction from Disparate Views using Neural Template Regularization},
  author = {Vora, Aditya and Patil, Akshay Gadi and Zhang, Hao},
  booktitle = {Neural Information Processing Systems},
  volume = {36},
  year = {2023},
  journal = {Neural Information Processing Systems},
  pages = {66768-66781},
  doi = {10.52202/075280-2915},
  publisher = {Neural Information Processing Systems Foundation, Inc. (NeurIPS)},
}

@inproceedings{wang2024unisdf,
  title = {UniSDF: Unifying Neural Representations for High-Fidelity 3D Reconstruction of Complex Scenes with Reflections},
  author = {Wang, Fangjinhua and Rakotosaona, Marie-Julie and Niemeyer, Michael and Szeliski, Richard and Pollefeys, Marc and Tombari, Federico},
  booktitle = {Neural Information Processing Systems},
  volume = {37},
  year = {2023},
  journal = {Neural Information Processing Systems},
  doi = {10.48550/arXiv.2312.13285},
}

@inproceedings{rojas2025hamst3r,
  title = {HAMSt3R: Human-Aware Multi-view Stereo 3D Reconstruction},
  author = {Rojas, Sara and Armando, M. and Ghamen, Bernard and Weinzaepfel, Philippe and Leroy, Vincent and Rogez, Grégory},
  booktitle = {IEEE International Conference on Computer Vision},
  pages = {5027--5037},
  year = {2025},
  journal = {IEEE International Conference on Computer Vision},
  doi = {10.48550/arXiv.2508.16433},
  publisher = {IEEE},
}

@article{kerbl20233dgaussians,
  title = {3D Gaussian Splatting for Real-Time Radiance Field Rendering},
  author = {Kerbl, Bernhard and Kopanas, Georgios and Leimkuehler, Thomas and Drettakis, G.},
  journal = {ACM Transactions on Graphics},
  volume = {42},
  number = {4},
  year = {2023},
  pages = {1-14},
  doi = {10.1145/3592433},
  publisher = {Association for Computing Machinery (ACM)},
}

@article{xu2025seres,
  title = {SERES: Semantic-aware neural reconstruction from sparse views},
  author = {Xu, Bo and Guo, Yuhu and Wang, Yuchao and Wang, Wenting and Yam, Yeung and Wang, C. C. and Le, Xinyi},
  journal = {IEEE Transactions on Visualization and Computer Graphics},
  year = {2025},
  doi = {10.1109/TVCG.2025.3619144},
}

@inproceedings{yuan2024sdmvs,
  title = {SD-MVS: Segmentation-Driven Deformation Multi-View Stereo with Spherical Refinement and EM Optimization},
  author = {Yuan, Z. and Cao, J. and Li, Z. and Jiang, H. and Wang, Z.},
  booktitle = {AAAI Conference on Artificial Intelligence},
  volume = {38},
  pages = {6871--6880},
  year = {2024},
  journal = {AAAI Conference on Artificial Intelligence},
  doi = {10.48550/arXiv.2401.06385},
  publisher = {Association for the Advancement of Artificial Intelligence (AAAI)},
}

@inproceedings{yang2024depthanythingv2,
  title = {Depth Anything V2},
  author = {Yang, Lihe and Kang, Bingyi and Huang, Zilong and Zhao, Zhen and Xu, Xiaogang and Feng, Jiashi and Zhao, Hengshuang},
  booktitle = {Neural Information Processing Systems},
  volume = {37},
  year = {2024},
  journal = {Neural Information Processing Systems},
  pages = {21875-21911},
  doi = {10.48550/arXiv.2406.09414},
  publisher = {Neural Information Processing Systems Foundation, Inc. (NeurIPS)},
}

@inproceedings{sparsevggt2026,
  title = {Faster VGGT with Block-Sparse Global Attention},
  author = {Wang, Chung-Shien Brian and Schmidt, Christian and Piekenbrinck, Jens and Leibe, Bastian},
  booktitle = {arXiv.org},
  year = {2025},
  note = {GitHub repository: brianwang00001/sparse-vggt},
  journal = {arXiv.org},
  doi = {10.48550/arXiv.2509.07120},
}

@article{speed3r2026,
  title = {Speed3R: Sparse Feed-forward 3D Reconstruction Models},
  author = {Ren, Weining and Tan, Xiao and Han, Kai},
  journal = {arXiv preprint arXiv:2603.08055},
  year = {2026},
}

@article{fastvggt2025,
  title = {FastVGGT: Training-free acceleration of visual geometry transformer},
  author = {Shen, You and Zhang, Zhipeng and Qu, Yansong and Cao, Liujuan},
  journal = {arXiv.org},
  year = {2025},
  doi = {10.48550/arXiv.2509.02560},
}

@article{flashvggt2025,
  title = {FlashVGGT: Efficient and Scalable Visual Geometry Transformers with Compressed Descriptor Attention},
  author = {Wang, Zipeng and Xu, Dan},
  journal = {arXiv.org},
  year = {2025},
  doi = {10.48550/arXiv.2512.01540},
}

@inproceedings{zpressor2026,
  title = {ZPressor: Bottleneck-Aware Compression for Scalable Feed-Forward 3DGS},
  author = {Wang, Weijie and Chen, Donny Y. and Zhang, Zeyu and Shi, Duochao and Liu, Akide and Zhuang, Bohan},
  booktitle = {arXiv.org},
  year = {2025},
  url = {https://openreview.net/forum?id=zbucdbZ0fU},
  journal = {arXiv.org},
  doi = {10.48550/arXiv.2505.23734},
}

\clearpage
\appendix

\section{Sparse Linear Attention (SLA) summary}
\label{app:sla-summary}

Sparse Linear Attention (SLA) is the lightweight attention module used to construct the Lite3R student. As described in the main method, SLA replaces dense self-attention with a hybrid module that combines a sparse geometric branch and a low-cost linear-context branch. Given input tokens $X\in\mathbb{R}^{N\times d}$ with projections $Q=XW_Q$, $K=XW_K$, and $V=XW_V$, Lite3R uses
\begin{equation}
A_{\mathrm{SLA}}(Q,K,V)=A_{\mathrm{sparse}}(Q,K,V)+\operatorname{Proj}\big(A_{\mathrm{lin}}(Q,K,V)\big).
\end{equation}
Here, the sparse branch preserves a small set of high-value query--key interactions that carry cross-view geometric correspondences, while the linear branch supplies low-cost global context. This replacement lowers token-mixing cost while maintaining a reasonable approximation to dense multi-view interaction.

\begin{algorithm}
\caption{Sparse Linear Attention (SLA)}
\label{alg:sla}
\begin{algorithmic}[1]
\Require Input tokens $\mathbf{X} \in \mathbb{R}^{N \times d}$
\Require Frozen projection matrices $\mathbf{W}_Q, \mathbf{W}_K, \mathbf{W}_V \in \mathbb{R}^{d \times d}$
\Require Trainable linear output projection $\mathbf{W}_O \in \mathbb{R}^{d \times d}$
\Require Sparse keep ratio $\lambda \in [0,1]$
\Ensure Output tokens $\mathbf{O} \in \mathbb{R}^{N \times d}$
\State $\mathbf{Q} \gets \mathbf{X}\mathbf{W}_Q$, $\mathbf{K} \gets \mathbf{X}\mathbf{W}_K$, $\mathbf{V} \gets \mathbf{X}\mathbf{W}_V$ \Comment{Frozen teacher-inherited projections}
\State $\mathbf{S} \gets \mathbf{Q}\mathbf{K}^{\top} / \sqrt{d}$ \Comment{Pairwise affinity scores}
\State $\mathcal{M} \gets \operatorname{TopKMask}(\mathbf{S}, \lambda)$ \Comment{Keep top-$\lambda$ fraction per query}
\State $\mathbf{A}_{\mathrm{sparse}} \gets \operatorname{Softmax}(\mathbf{S} \odot \mathcal{M})$
\State $\mathbf{O}_{\mathrm{sparse}} \gets \mathbf{A}_{\mathrm{sparse}}\mathbf{V}$ \Comment{Sparse geometric branch}
\State $\phi(\mathbf{Q}) \gets \operatorname{ELU}(\mathbf{Q}) + 1$, $\phi(\mathbf{K}) \gets \operatorname{ELU}(\mathbf{K}) + 1$ \Comment{Linear branch uses frozen $\mathbf{Q},\mathbf{K}$ features}
\State $\mathbf{T} \gets \phi(\mathbf{K})^{\top}\mathbf{V}$ \Comment{Linear key--value summary}
\State $\mathbf{z} \gets \phi(\mathbf{K})^{\top}\mathbf{1}_N$ \Comment{Linear normalization term}
\State $\mathbf{O}_{\mathrm{lin}} \gets (\phi(\mathbf{Q})\mathbf{T}) \oslash (\phi(\mathbf{Q})\mathbf{z})$ \Comment{Linear-context branch}
\State $\mathbf{O} \gets \mathbf{O}_{\mathrm{sparse}} + \mathbf{O}_{\mathrm{lin}}\mathbf{W}_O$ \Comment{Only $\mathbf{W}_O$ is trainable during adaptation}
\State \Return $\mathbf{O}$
\end{algorithmic}
\end{algorithm}

\section{FP8-Aware Quantization-Aware Training}
\label{appendix:fp8_qat}

This section provides technical details of our FP8-aware Quantization-Aware Training (QAT) approach, which enables efficient deployment of large-scale vision-language models while maintaining reconstruction quality. Our method is model-agnostic and can be applied to various transformer-based architectures.

\subsection{FP8 E4M3 Format}

We adopt the FP8 E4M3 format (\texttt{float8\_e4m3fn}) for quantization, which allocates 1 sign bit, 4 exponent bits, and 3 mantissa bits. This format provides a dynamic range of approximately $[-448, 448]$ with reduced precision compared to higher-precision floating-point formats (e.g., FP32, FP16, BF16), making it suitable for efficient inference on modern accelerators with native FP8 support.

\subsection{Scaled FP8 Quantization with Straight-Through Estimator}

Our FP8 fake-quantization simulates the quantization noise during training while maintaining full-precision gradients through a straight-through estimator (STE). For a tensor $\mathbf{x} \in \mathbb{R}^{d}$, the scaled FP8 quantization is defined as:

\begin{equation}
\text{scale} = \frac{\max(|\mathbf{x}|)}{\text{FP8\_MAX}}, \quad \text{FP8\_MAX} = 448
\end{equation}

\begin{equation}
\mathbf{x}_{\text{scaled}} = \text{clamp}\left(\frac{\mathbf{x}}{\text{scale}}, \text{FP8\_MIN}, \text{FP8\_MAX}\right)
\end{equation}

\begin{equation}
\mathbf{x}_{\text{quant}} = \text{FP8}(\mathbf{x}_{\text{scaled}}) \cdot \text{scale}
\end{equation}

\begin{equation}
\mathbf{x}_{\text{STE}} = \mathbf{x} + (\mathbf{x}_{\text{quant}} - \mathbf{x}).detach()
\end{equation}

where $\text{FP8}(\cdot)$ denotes casting to FP8 E4M3 format, and $.detach()$ stops gradient flow. The STE ensures that gradients flow through as if no quantization occurred: $\frac{\partial \mathbf{x}_{\text{STE}}}{\partial \mathbf{x}} = \mathbf{I}$.

\subsection{Per-Tensor Dynamic Scaling}

We employ different scaling strategies for weights and activations to preserve their respective dynamic ranges:

\begin{itemize}
    \item \textbf{Weight quantization}: Per-output-channel scaling. For weight matrix $\mathbf{W} \in \mathbb{R}^{d_{\text{out}} \times d_{\text{in}}}$, we compute scale factors per row:
    \begin{equation}
    \text{scale}_i = \frac{\max_j |\mathbf{W}_{i,j}|}{\text{FP8\_MAX}}, \quad i = 1, \ldots, d_{\text{out}}
    \end{equation}

    \item \textbf{Activation quantization}: Per-token dynamic scaling. For activation tensor $\mathbf{X} \in \mathbb{R}^{N \times d}$ with $N$ tokens, we compute scale factors per token:
    \begin{equation}
    \text{scale}_i = \frac{\max_j |\mathbf{X}_{i,j}|}{\text{FP8\_MAX}}, \quad i = 1, \ldots, N
    \end{equation}
\end{itemize}

This granular scaling strategy minimizes quantization error by adapting to the local magnitude distribution of each tensor dimension.

\subsection{FP8 Fake-Quantization Linear Layer}

Algorithm~\ref{alg:fp8_linear} describes our FP8 fake-quantization linear layer, which wraps a standard linear layer with FP8 quantization simulation. This module can replace any \texttt{nn.Linear} layer in transformer-based architectures.

\begin{algorithm}[t]
\caption{FP8 Fake-Quantization Linear Layer}
\label{alg:fp8_linear}
\begin{algorithmic}[1]
\Require Input $\mathbf{x} \in \mathbb{R}^{N \times d_{\text{in}}}$, weight $\mathbf{W} \in \mathbb{R}^{d_{\text{out}} \times d_{\text{in}}}$, bias $\mathbf{b} \in \mathbb{R}^{d_{\text{out}}}$
\Require FP8 dtype: \texttt{float8\_e4m3fn}, enable activation quantization flag
\Ensure Output $\mathbf{y} \in \mathbb{R}^{N \times d_{\text{out}}}$
\State
\State \textbf{Activation Quantization (Per-Token):}
\If{enable\_act\_quant}
    \State $\text{scale}_{\mathbf{x}} \gets \max(|\mathbf{x}|, \text{dim}=-1, \text{keepdim}=\text{True}) / \text{FP8\_MAX}$
    \State $\mathbf{x}_{\text{scaled}} \gets \text{clamp}(\mathbf{x} / \text{scale}_{\mathbf{x}}, \text{FP8\_MIN}, \text{FP8\_MAX})$
    \State $\mathbf{x}_{\text{fp8}} \gets \text{FP8}(\mathbf{x}_{\text{scaled}}) \cdot \text{scale}_{\mathbf{x}}$
    \State $\mathbf{x}_q \gets \mathbf{x} + (\mathbf{x}_{\text{fp8}} - \mathbf{x}).detach()$ \Comment{STE}
\Else
    \State $\mathbf{x}_q \gets \mathbf{x}$
\EndIf
\State
\State \textbf{Weight Quantization (Per-Output-Channel):}
\State $\text{scale}_{\mathbf{W}} \gets \max(|\mathbf{W}|, \text{dim}=-1, \text{keepdim}=\text{True}) / \text{FP8\_MAX}$
\State $\mathbf{W}_{\text{scaled}} \gets \text{clamp}(\mathbf{W} / \text{scale}_{\mathbf{W}}, \text{FP8\_MIN}, \text{FP8\_MAX})$
\State $\mathbf{W}_{\text{fp8}} \gets \text{FP8}(\mathbf{W}_{\text{scaled}}) \cdot \text{scale}_{\mathbf{W}}$
\State $\mathbf{W}_q \gets \mathbf{W} + (\mathbf{W}_{\text{fp8}} - \mathbf{W}).detach()$ \Comment{STE}
\State
\State \textbf{Linear Transformation:}
\State $\mathbf{y} \gets \mathbf{x}_q \mathbf{W}_q^\top + \mathbf{b}$
\State
\State \Return $\mathbf{y}$
\end{algorithmic}
\end{algorithm}

\subsection{Training Procedure}

Our FP8-aware QAT can be integrated into existing training pipelines with minimal modifications. The general procedure consists of:

\begin{enumerate}
    \item \textbf{Baseline Training}. Train or fine-tune the model with its original precision (typically FP32, FP16, or BF16) until convergence. This establishes a strong baseline and ensures the model has learned the task-specific representations.

    \item \textbf{FP8 QAT Fine-tuning}. Replace all target linear layers with FP8 fake-quantization layers and continue training for a small number of epochs (typically 1-5). During this stage:
    \begin{itemize}
        \item Forward pass: Both weights and activations are quantized to FP8 E4M3 with dynamic per-tensor scaling
        \item Backward pass: Gradients flow through the STE as if no quantization occurred, maintaining the original precision for gradient computation
        \item Learning rate: Typically reduced by 10$\times$ compared to the baseline training phase
        \item Trainable parameters: Can be all parameters or a subset (e.g., task-specific heads, adapter layers)
    \end{itemize}
\end{enumerate}

This two-stage approach allows the model to adapt to quantization noise while leveraging the representations learned during higher-precision training.

\subsection{Implementation Details}

\textbf{Selective quantization}: Our framework supports flexible quantization policies. Users can specify which layers to quantize based on module names or types. Common strategies include:
\begin{itemize}
    \item Quantize all linear layers in the model
    \item Quantize only attention and MLP layers, keeping normalization and embedding layers in full precision
    \item Skip quantization for small layers (e.g., projection heads with $<$1M parameters)
\end{itemize}

\textbf{Numerical stability}: We add a small epsilon ($\epsilon = 10^{-8}$) when computing scale factors to prevent division by zero for near-zero tensors. Additionally, we clamp the scaled values to the valid FP8 range before casting.

\textbf{Training efficiency}: The FP8 fake-quantization adds minimal overhead during training ($<5\%$ slowdown) since the quantization operations are implemented as efficient CUDA kernels, and the STE requires no additional backward computation beyond the standard autograd graph.

\subsection{Deployment Considerations}

At inference time, the FP8-quantized model can be deployed on hardware accelerators with native FP8 support (e.g., NVIDIA H100, AMD MI300). The per-tensor scaling factors are stored alongside the quantized weights, enabling efficient dequantization during matrix multiplication. Our approach achieves:

\begin{itemize}
    \item \textbf{Memory reduction}: FP8 weights occupy half the memory of FP16 weights, or one-quarter the memory of FP32 weights, enabling larger batch sizes or longer sequences
    \item \textbf{Minimal accuracy degradation}: Empirically, FP8 QAT maintains task performance within 1-2\% of the higher-precision baseline across various vision-language tasks
    \item \textbf{Hardware efficiency}: Native FP8 tensor cores provide up to 2$\times$ throughput compared to FP16 on supported hardware, translating to faster inference and higher throughput
\end{itemize}

\textbf{Compatibility}: For hardware without native FP8 support, the quantized model can be deployed using simulated FP8 arithmetic (dequantize to the original precision, compute, quantize back), though this sacrifices the speed benefits while retaining the memory advantages.

\section{Supplementary Efficiency and Sensitivity Analysis}
\label{appendix:main-results-comprehensive}

\begin{figure*}[t]
  \centering
  \includegraphics[width=0.98\textwidth]{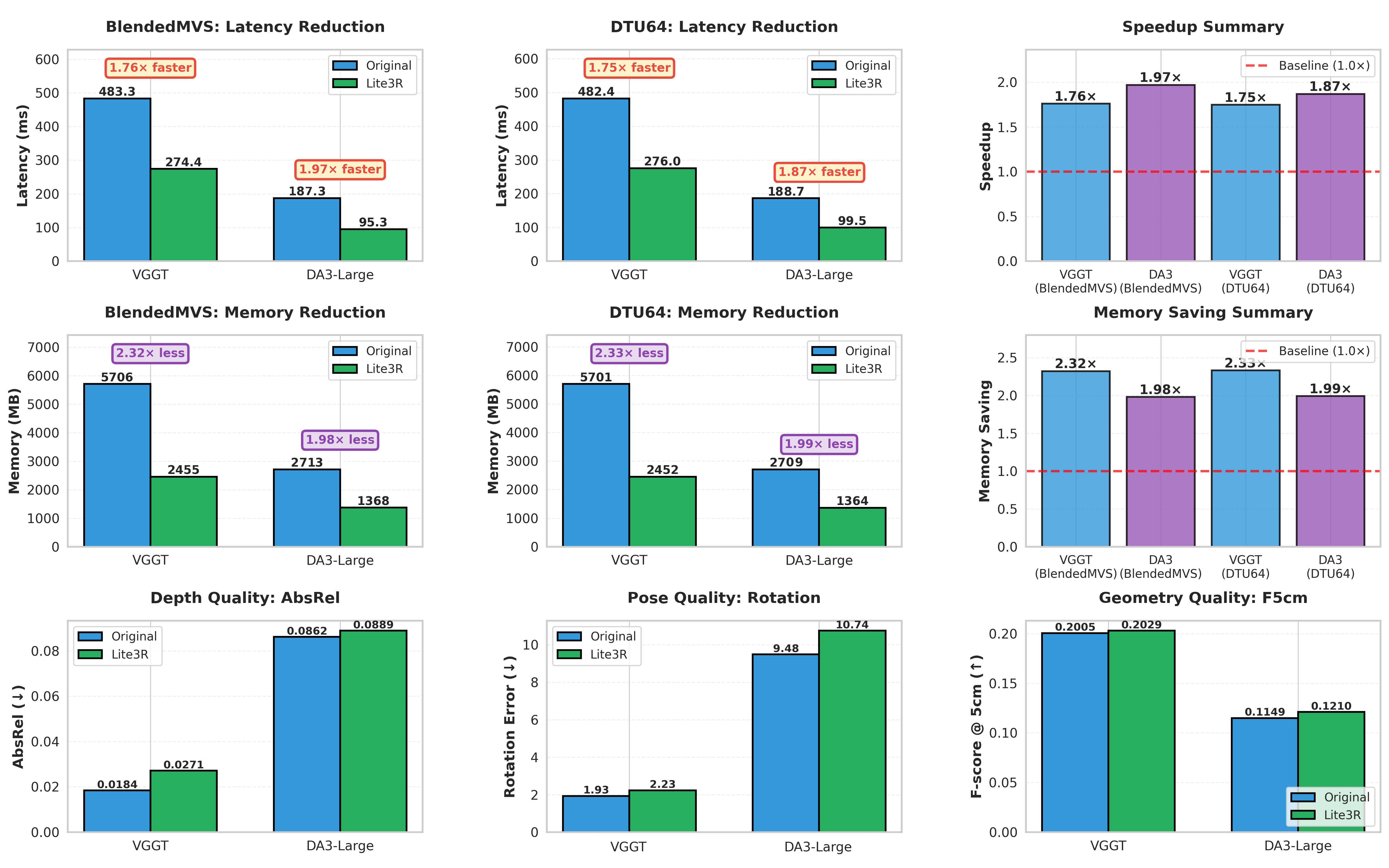}
  \caption{\textbf{Comprehensive visualization of Lite3R main results across nine experimental settings.} The bar charts summarize the key quality and efficiency metrics reported in the main paper, highlighting how Lite3R compares with the corresponding higher-precision baselines across different backbones, datasets, and evaluation dimensions.}
  \label{fig:main-results-comprehensive}
\end{figure*}

This section supplements the main experiments with a broader view of the trends summarized by Figure~\ref{fig:main-results-comprehensive}. While the main paper emphasizes compact tables, the appendix visualization helps compare Lite3R across backbones, datasets, and metric groups in a more global way.

For DA3-Large, the parameter allocation highlights the extreme efficiency of the adaptation strategy after SLA replacement. The full model contains 411.06M parameters, but only 0.11M parameters (0.03\%) remain trainable, while 410.94M parameters (99.97\%) are frozen. This comes from replacing 28 attention modules with SLA and updating only the corresponding \texttt{proj\_lin} layers, each of which contains 4,096 trainable parameters. Most parameters still reside in the DinoV2 backbone (304.47M), followed by the camera encoder (50.94M), DPT head (47.23M), and camera decoder (8.41M), but the head and decoder are fully frozen. Compared with VGGT, whose trainable ratio is 3.1\%, DA3-Large uses an even more parameter-efficient adaptation regime. This helps explain its result pattern: adaptation capacity is concentrated in a tiny set of projection layers, so the model still gains strong latency and memory benefits but has less flexibility than VGGT under structural and numerical changes.

\paragraph{Layer-wise sensitivity score.}
For each linear layer with weight tensor $\mathbf{W}$, we compute a quantization sensitivity score that combines three statistical indicators of vulnerability to FP8 perturbation:
\begin{equation}
S(\mathbf{W}) = 0.4 \cdot \frac{\|\mathbf{W}\|_\infty}{10} + 0.3 \cdot r_{\mathrm{out}}(\mathbf{W}) + 0.3 \cdot \frac{\mathrm{kurt}(\mathbf{W})}{10},
\end{equation}
where $\|\mathbf{W}\|_\infty$ is the dynamic range, $r_{\mathrm{out}}(\mathbf{W})$ is the fraction of weights beyond $3\sigma$, and $\mathrm{kurt}(\mathbf{W})$ is the kurtosis. Larger dynamic range, more outliers, and heavier tails all indicate higher sensitivity to FP8 quantization.

\begin{figure*}[t]
  \centering
  \includegraphics[width=0.98\textwidth]{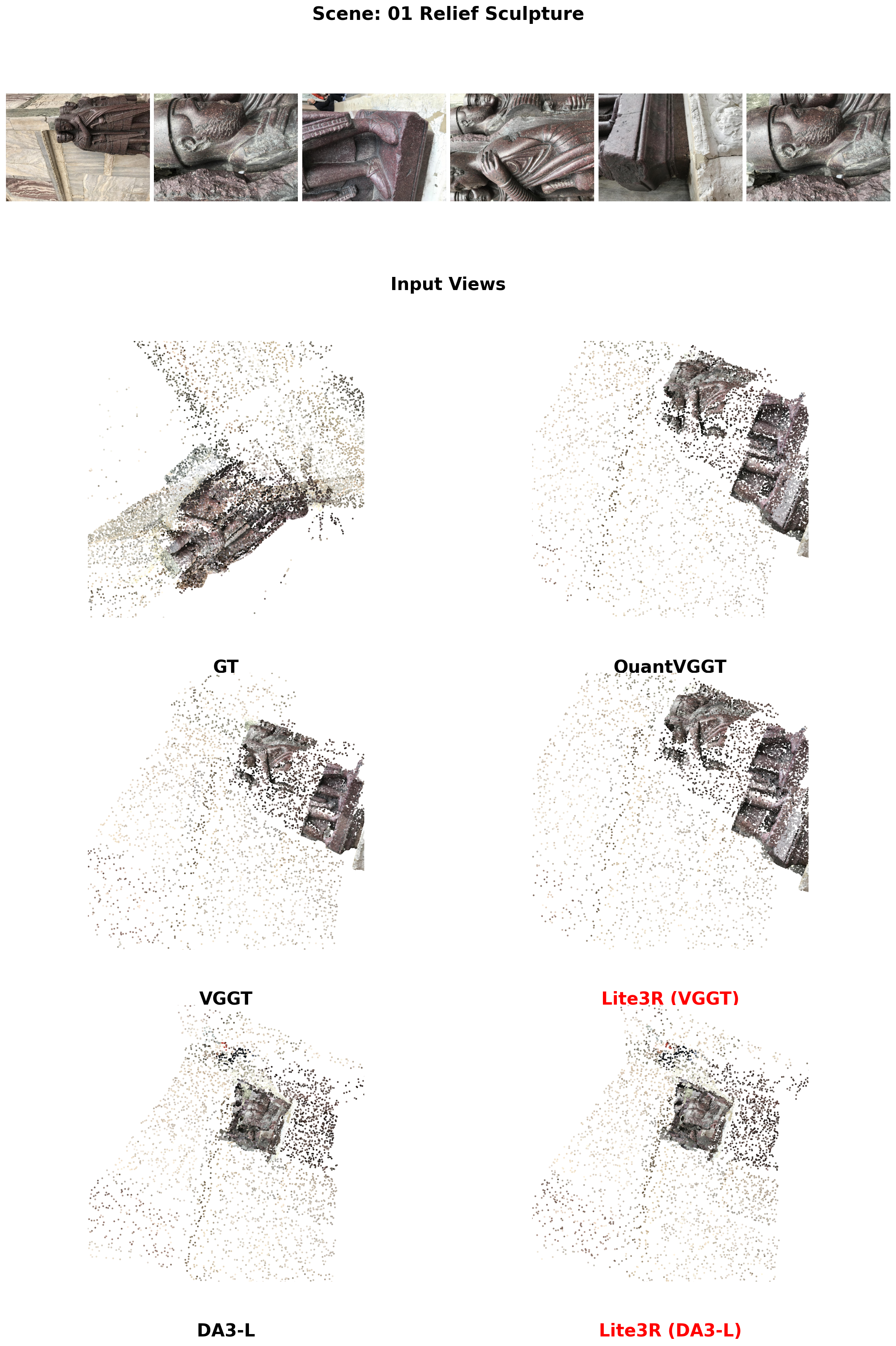}
  \label{v1}
\end{figure*}

\begin{figure*}[t]
  \centering
  \includegraphics[width=0.98\textwidth]{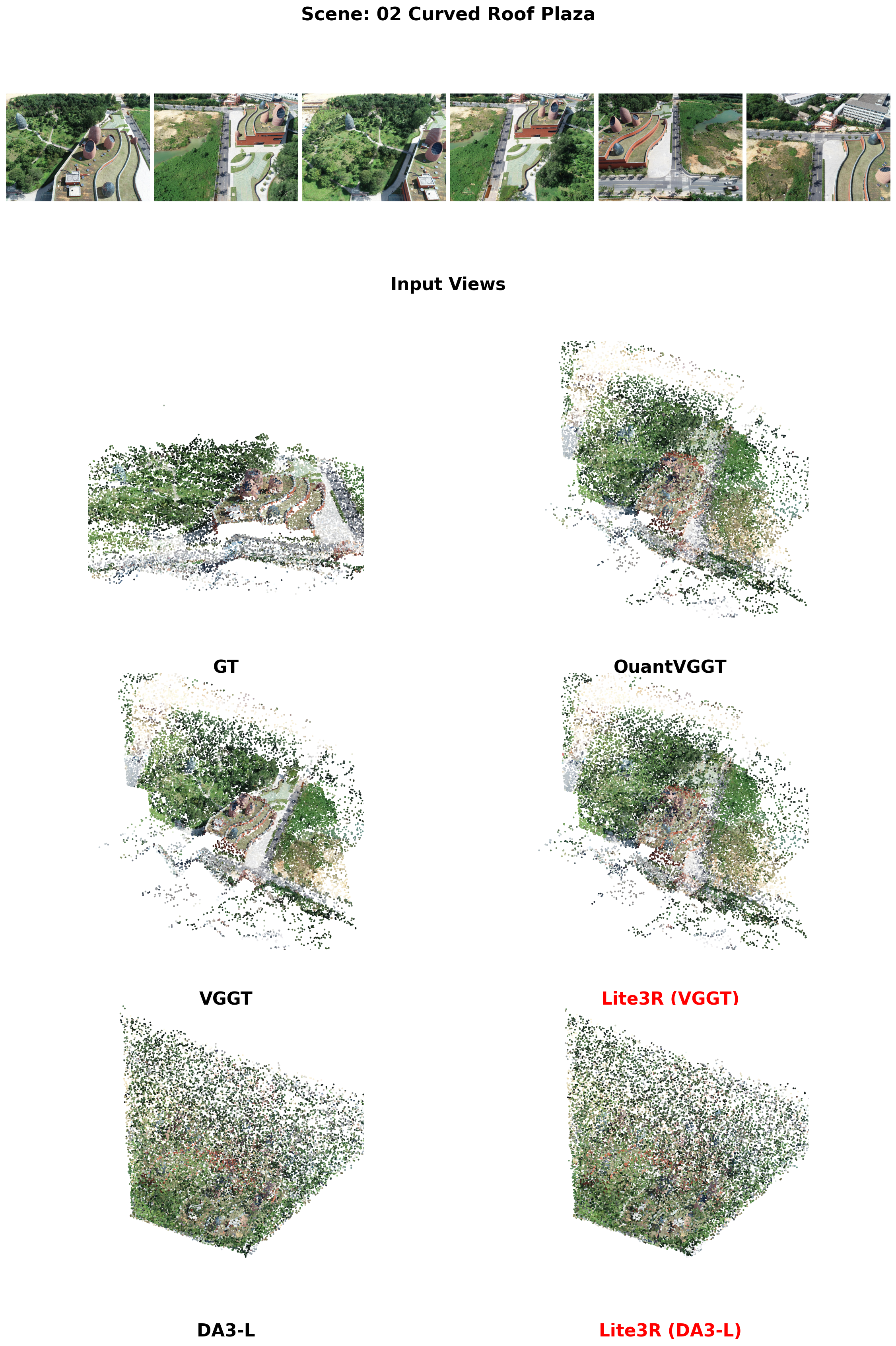}
  \label{v2}
\end{figure*}

\begin{figure*}[t]
  \centering
  \includegraphics[width=0.98\textwidth]{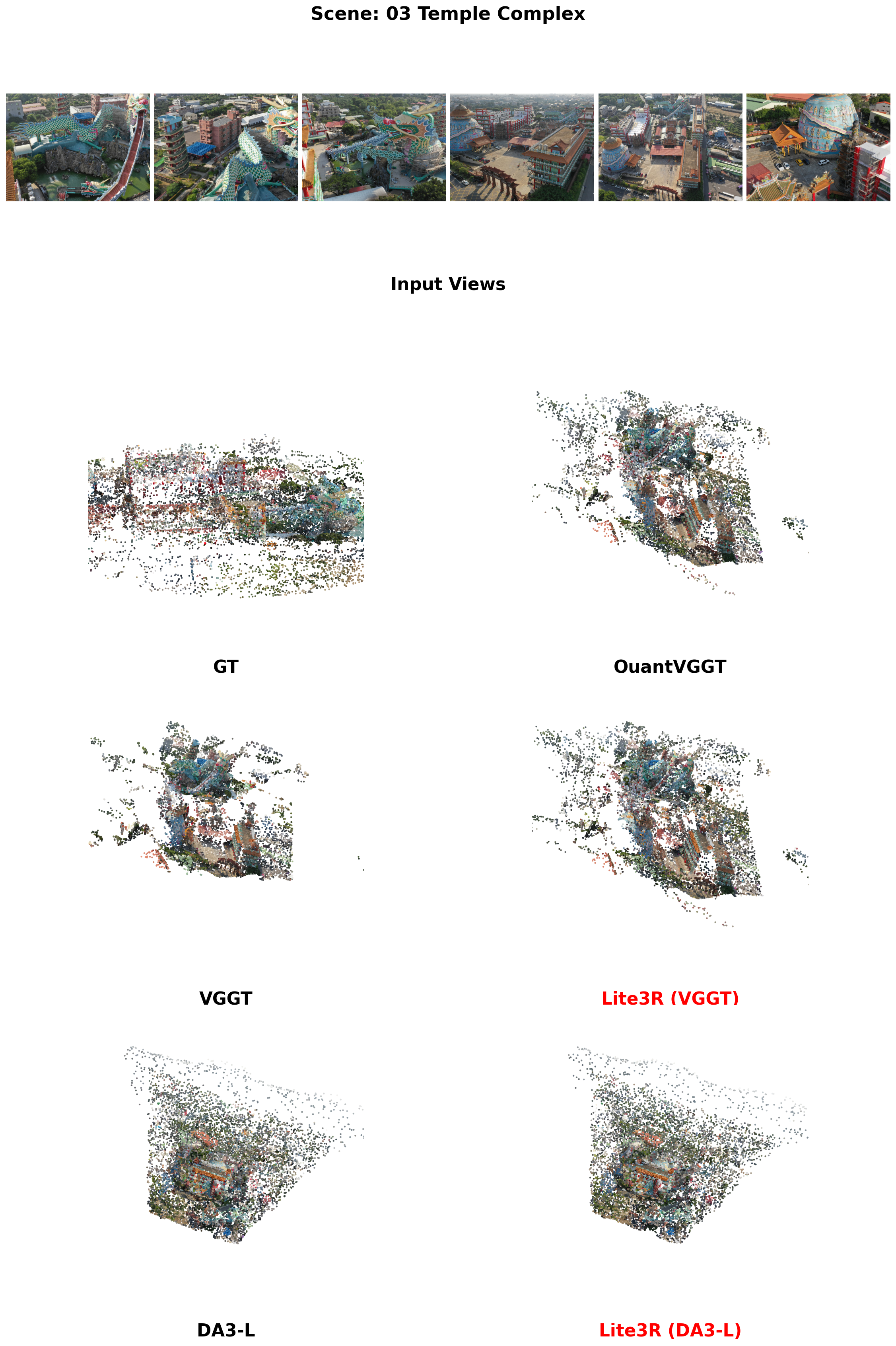}
  \label{v3}
\end{figure*}

\begin{figure*}[t]
  \centering
  \includegraphics[width=0.98\textwidth]{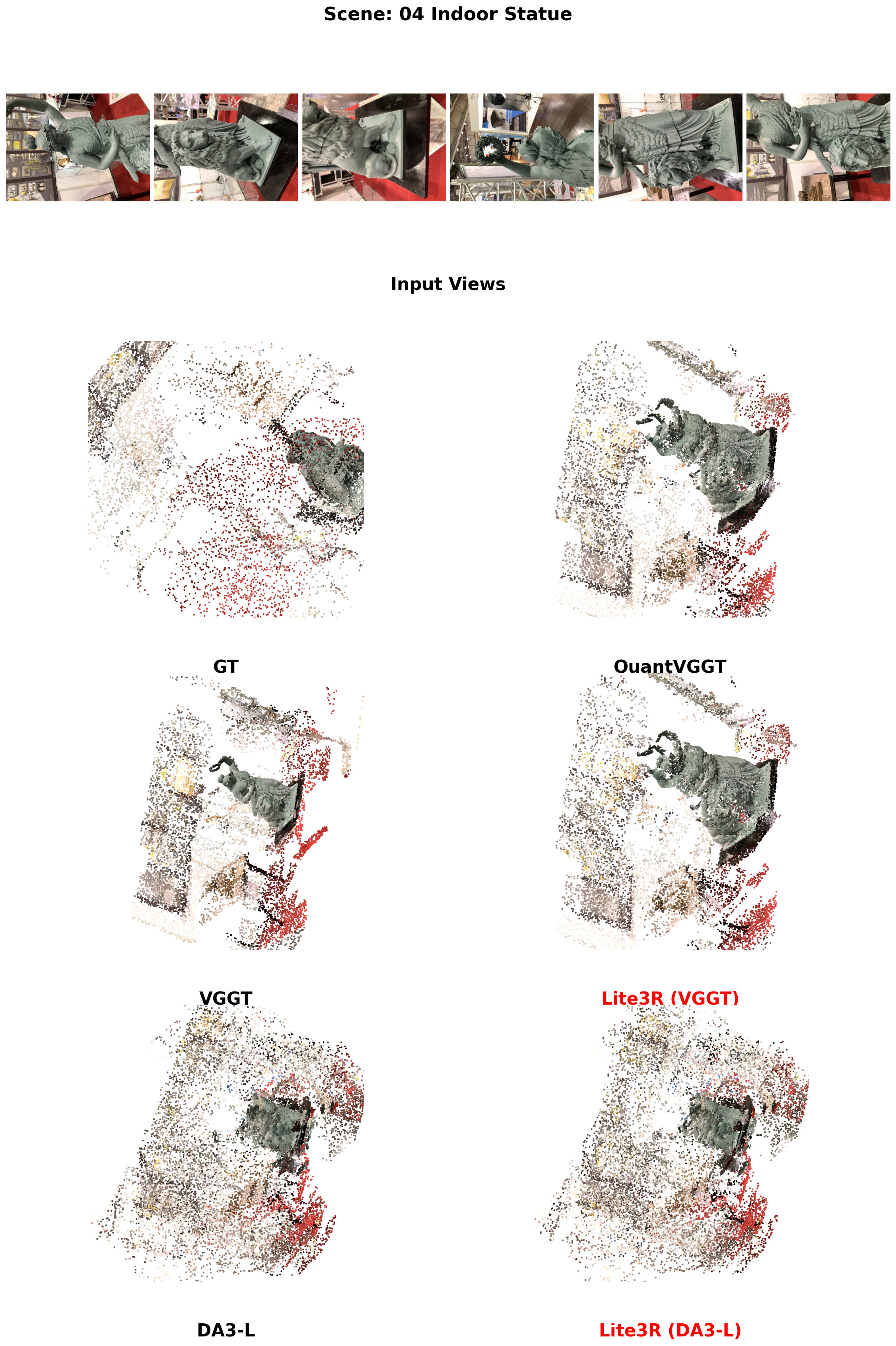}
  \label{v4}
\end{figure*}

\begin{figure*}[t]
  \centering
  \includegraphics[width=0.98\textwidth]{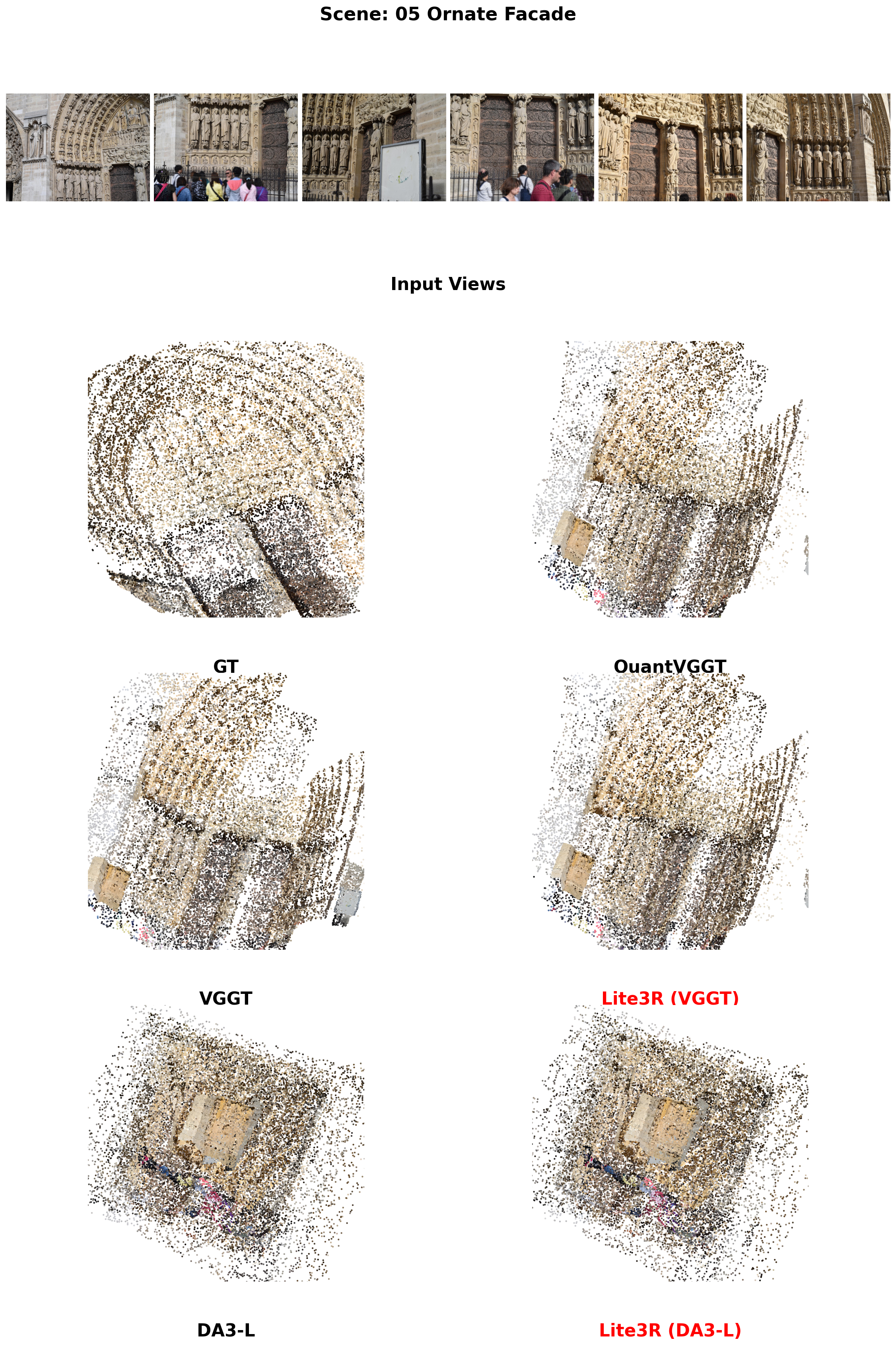}
  \label{v5}
\end{figure*}

\begin{figure*}[t]
  \centering
  \includegraphics[width=0.98\textwidth]{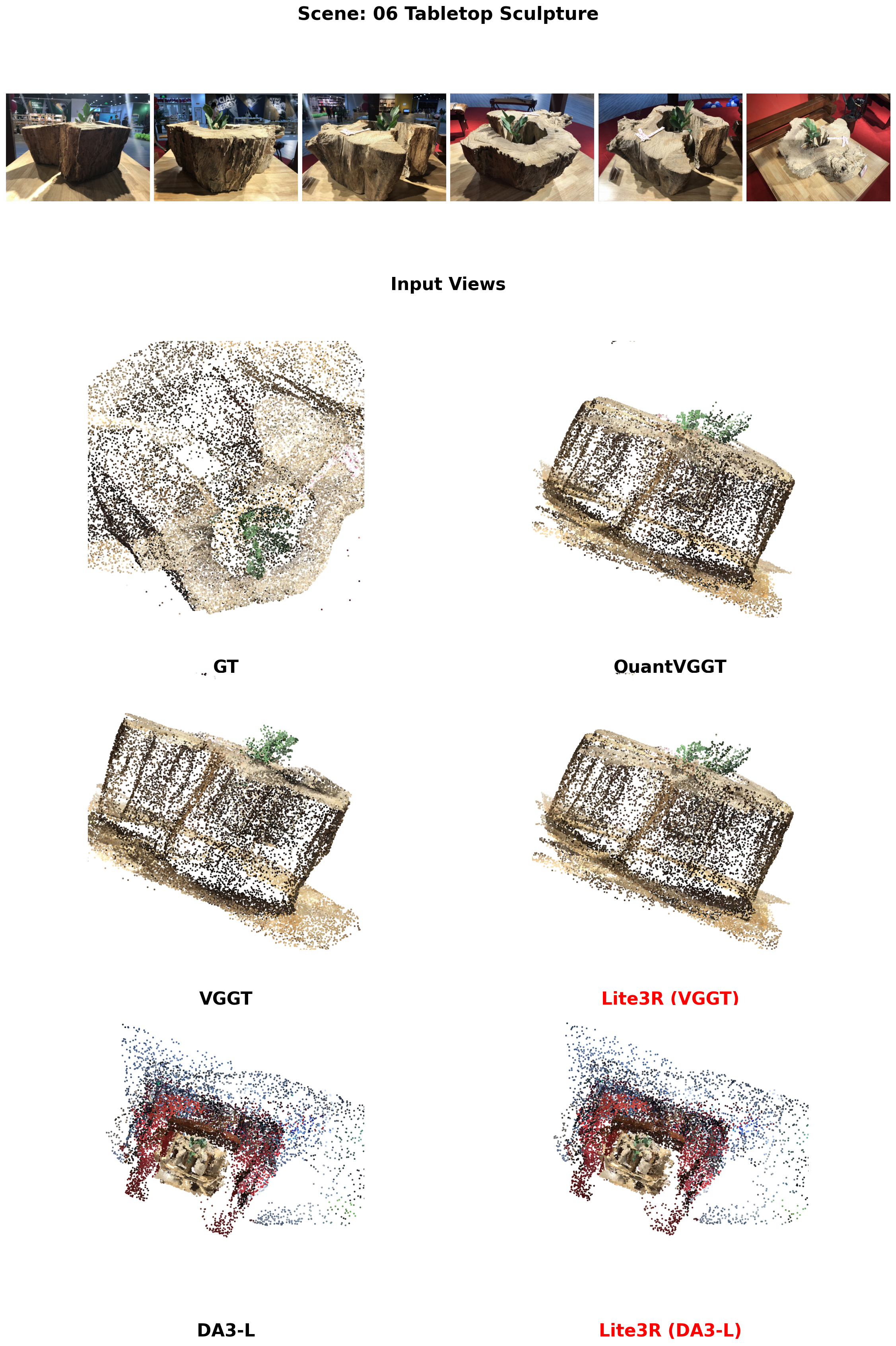}
  \label{v6}
\end{figure*}

\begin{figure*}[t]
  \centering
  \includegraphics[width=0.98\textwidth]{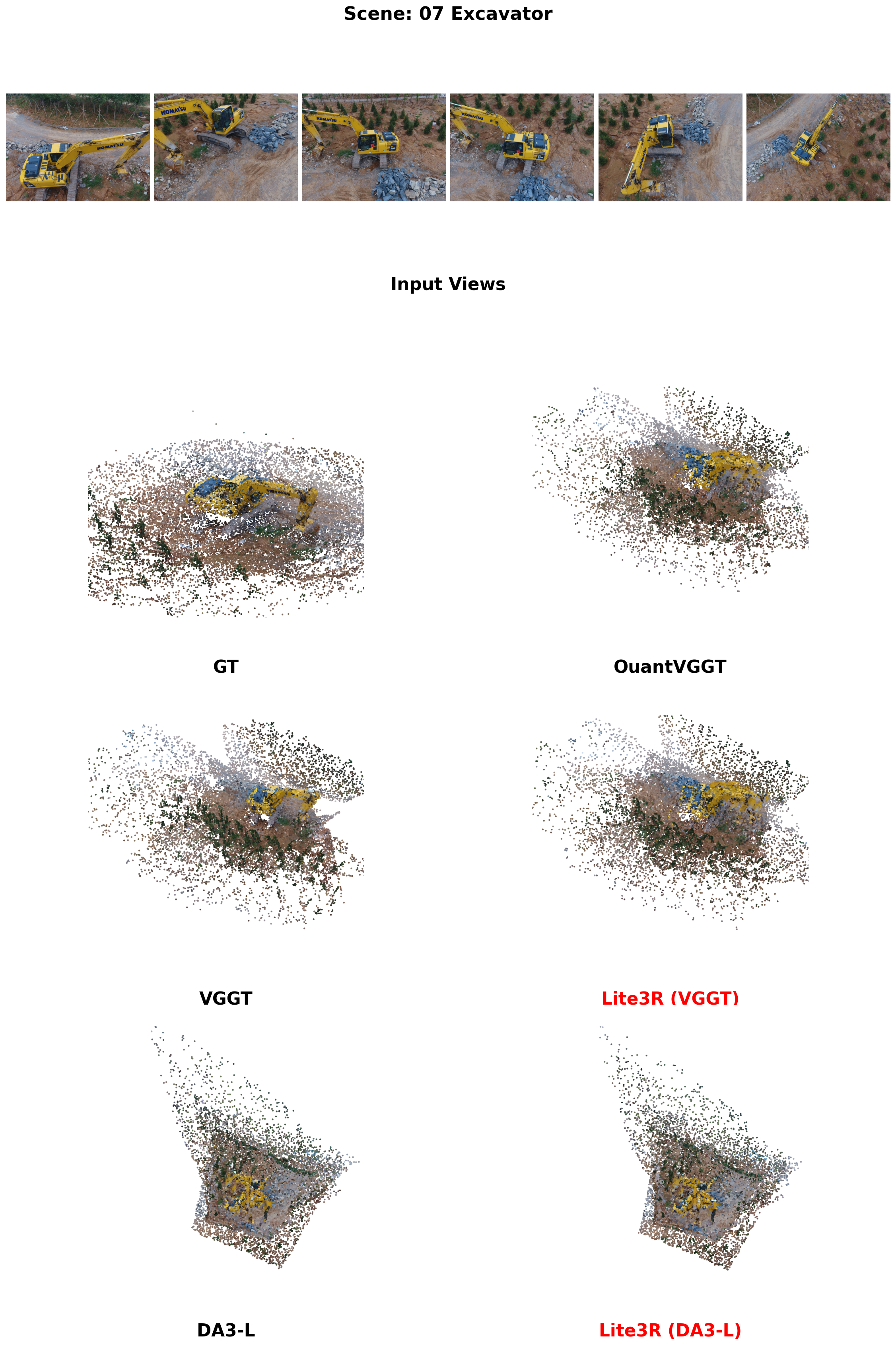}
  \label{v7}
\end{figure*}

\begin{figure*}[t]
  \centering
  \includegraphics[width=0.98\textwidth]{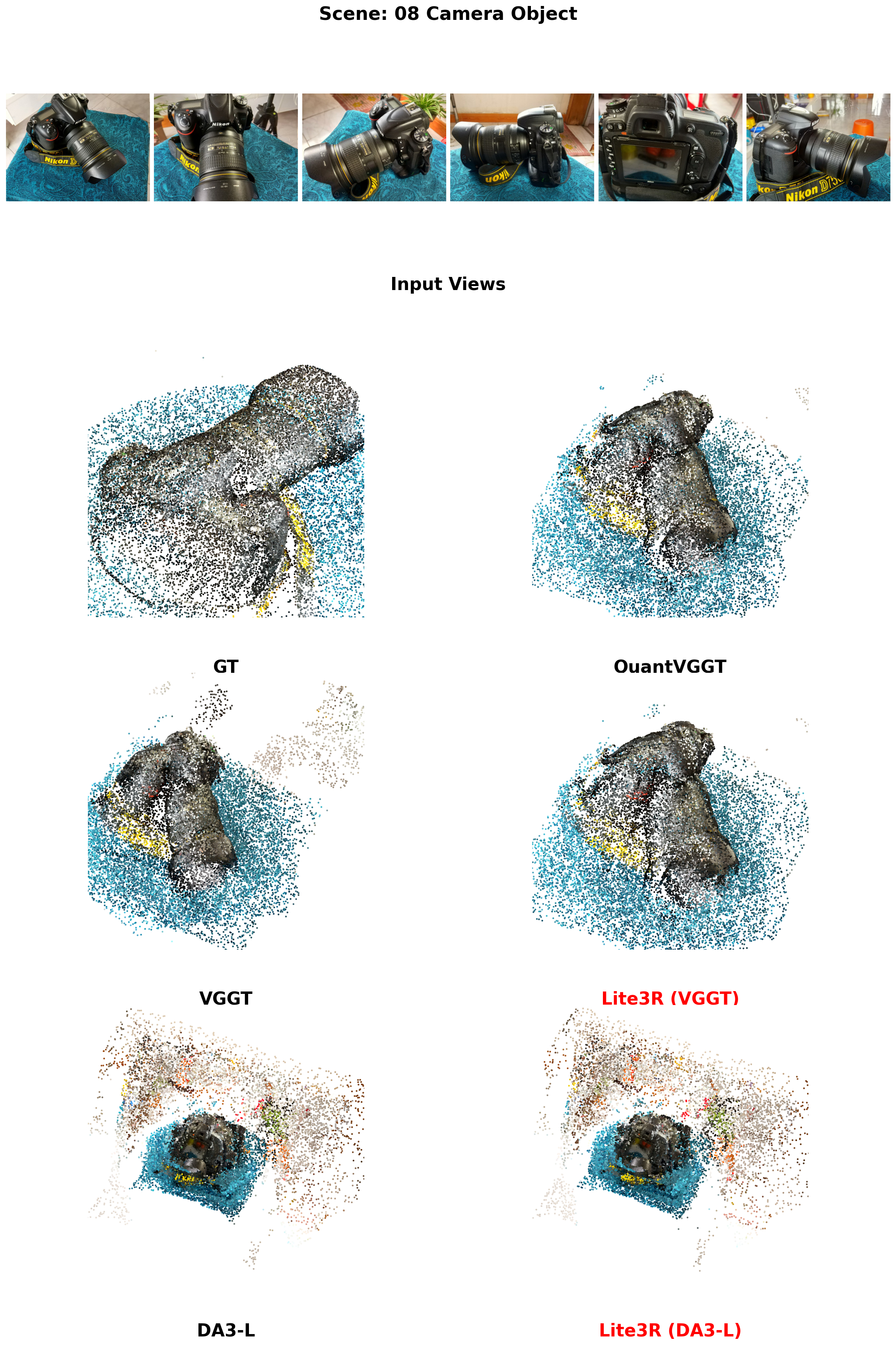}
  \label{v8}
\end{figure*}

\end{document}